\documentclass[lettersize,journal]{IEEEtran}
\newenvironment{proof}{{\indent \indent \it Proof:\quad}}{\hfill $\blacksquare$\par}
\usepackage{graphicx} 
\usepackage{subfigure}
\usepackage{amsmath,amsfonts}
\usepackage{algorithmic}
\usepackage{algorithm}
\usepackage{array}
\usepackage[caption=false,font=normalsize,labelfont=sf,textfont=sf]{subfig}
\usepackage{textcomp}
\usepackage{stfloats}
\usepackage{url}
\usepackage{verbatim}
\usepackage{graphicx}
\usepackage{cite}
\usepackage{amssymb}
\usepackage{booktabs}
\usepackage{threeparttable}
 \usepackage{tabularx}
\usepackage{multirow}
\usepackage{multicol}

\newtheorem{theorem}{Theorem}
\newtheorem{definition}{Definition}
\begin{document}

\title{Quaternion Optimized Model with Sparse Regularization for Color Image Recovery}
\author{Liqiao Yang, Yang Liu
	\IEEEmembership{Senior Member, IEEE}, Kit Ian Kou
	\thanks{Liqiao Yang is with the Department of Mathematics, Faculty of Science and Technology, University of Macau, Macau, China (e-mail: liqiaoyoung@163.com.}
	\thanks{Yang Liu is with the College of Mathematics and Computer Science and the College of Mathematical Medicine, Zhejiang Normal University, Jinhua 321004, China (e-mail: liuyang@zjnu.edu.cn).}
    \thanks{Kit Ian Kou is with the Department of Mathematics, Faculty of Science and Technology, University of Macau, Macau, China(e-mail: kikou@umac.mo)}}

\markboth{Journal of \LaTeX\ Class Files,~Vol.~14, No.~8, August~2021}%
{Shell \MakeLowercase{\textit{et al.}}: A Sample Article Using IEEEtran.cls for IEEE Journals}


\maketitle

\begin{abstract}
This paper addresses the color image completion problem in accordance with low-rank quatenrion matrix optimization that is characterized by sparse regularization in a transformed domain. This research was inspired by an appreciation of the fact that different signal types, including audio formats and images, possess structures that are inherently sparse in respect of their respective bases. Since color images can be processed as a whole in the quaternion domain, we depicted the sparsity of the color image in the quaternion discrete cosine transform (QDCT) domain. In addition, the representation of a low-rank structure that is intrinsic to the color image is a vital issue in the quaternion matrix completion problem. To achieve a more superior low-rank approximation, the quatenrion-based truncated nuclear norm (QTNN) is employed in the proposed model. Moreover, this model is facilitated by a competent alternating direction method of multipliers (ADMM) based on the algorithm. Extensive experimental results demonstrate that the proposed method can yield vastly superior completion performance in comparison with the state-of-the-art low-rank matrix/quaternion matrix approximation methods tested on color image recovery.
\end{abstract}


\section{Introduction}
\IEEEPARstart{I}{N} terms of image processing, the purpose of color image completion is to use the known pixels in an image, however limited, to recover missing pixels. To achieve this, many matrix-based completion (MC) approaches have been designed, though in  general, a color image is processed by separating RGB channels to three matrices and prior knowledge about the desired model is then used  to achieve inpainting. An effective and widely used  source of such prior knowledge is low-rankness.

In \cite{DBLP:journals/focm/CandesR09}, the nuclear norm (NN) was proved to be the tightest feasible convex relaxation of the matrix rank function, being NP hard to further minimize. Following on from this, other researchers have  shown that when the singular value is treated equally in the nuclear norm minimization process, the ensuing inpainting results are sub-optimal. Based on this, various work has been developed to optimize NN, such as the weighted nuclear norm \cite{DBLP:journals/ijcv/GuXMZFZ17},  the truncated nuclear norm (TNN) \cite{DBLP:journals/pami/HuZYLH13}, and the logarithmic norm \cite{DBLP:journals/tip/ChenJLZ21}, etc. However, although these approaches have improved the inpainting results for color images, especially as compared with classic NN methods, such matrix-based approaches involve dimension reduction, which can destroy the structure of the color image. 

To avoid this, quaternion-based approaches have gradually become more commonly used in image processing, as these allow the values of one pixel in the color image to be put in three imaginary parts of one quaternion to form a more reliable quaternion matrix. As the extension of a complex number, a quaternion number provides a representation that exactly matches the structure of the color pixel usually containing three values drawn from the RGB channels separately. Thus, quaternion-based approaches can be widely used in various types of color image processing, including color image recovery \cite{DBLP:journals/tip/ChenXZ20, DBLP:journals/tsp/MiaoK20, DBLP:journals/ijon/YuZY19}, color image watermarking \cite{DBLP:journals/jss/WangWYN13},  color face recognition \cite{DBLP:journals/tip/ZouKW16}, and so on. 
 
 For color image completion in the quaternion domain, the typical prior knowledge type is the low-rankness, analogous to the prior matrix-based cases. As an example, the authors in \cite{DBLP:journals/tip/ChenXZ20} proposed a low-rank quaternion approximation model based on several modified quaternion nuclear norm (QNN), including laplace, geman, and weighted Schatten-$\gamma$ functions. Offering two reasons why this series of methods could improve the inpainting performance. The first is that a single color image can be put in one quaternion matrix for processing, allowing correlation among RGB channels may be preserved, while the other is that these improved functions can approximate the rank of the quaternion matrix more precisely than QNN, with the latter being identical to matrix-based methods. However, calculating the quaternion singular value decomposition (QSVD) is computationally complex, with the operation needed to calculate the singular value decomposition of  corresponding complex matrices being twice the size of the original quaternion matrix in each iteration. 
 
 To overcome the time-consuming nature of QSVD,  researchers in \cite{DBLP:journals/tsp/MiaoK20}developed a  low-rank quaternion matrix factorization approach by factorizing the target quaternion matrix into the product of two smaller quaternion factor matrices. This factorization method is based on three kinds of quaternion bilinear matrix norms, quaternion double Frobenius norm (Q-DFN), quaternion double nuclear norm (Q-DNN), and quaternion Frobenius/nuclear norm (Q-FNN). In this factorization model, the low-rank estimation process retains higher levels of computing efficiency due to the fact that only two smaller quaternion factor matrices need to be optimized. However, as in the matrix-based cases, this factorization method may become trapped in the local minima \cite{DBLP:conf/iccv/CabralTCB13}, causing the computation of  QSVD for factor quaternion matrices to remain necessary in each iteration.
 
 As described above, the various low-rank completion methods can be divided into two branches: rank estimators supported by various regularization processes and low-rank factorization. Based on such assumptions about   low-rankness, the previously developed quaternion-based methods can thus process color images overall to obtain better recovery results;   however, they ignore other important properties such as sparsity. The direct motivation to reconsider this can be derived from the fact that various kinds of signals, including audio and images, have naturally sparse structures, with regard to given bases such as  Fourier and wavelet \cite{DBLP:journals/pieee/WrightMMSHY10}. This fact has inspired new approaches in signal processing in recent years, and especially for vision tasks based on matrix optimization, the $l_1$ norm has now been used extensively, especially for face reorganization \cite{DBLP:journals/jvcir/LuMGZL13, DBLP:journals/pami/WrightYGSM09 }, image denoising and completion \cite{DBLP:journals/spic/DongXGHW18, DBLP:conf/eccv/XuZZ18, DBLP:journals/pr/FanC17a, DBLP:conf/eccv/LiangRZM12}, and similar tasks. In terms of image completion, as  highlighted in \cite{DBLP:conf/eccv/LiangRZM12}, sparsity is also an important property that can be used in MC. Specifically, in certain transformed domains, piecewise smooth functions  have the property of being sparse and, based on this, in \cite{DBLP:journals/spic/DongXGHW18}, such sparsity is depicted by using the $l_1$ norm in the discrete cosine transform (DCT), alongside  low-rankness was as depicted by TNN. The TNN was found to obtain a more accurate approximation of rank than classic NN in this circumstance, as the first few largest singular values, despite being largest, did not influence the rank. Moreover, the TNN was also found to reduce computation time for the SVD of the target matrix ($\mathbf{X}\in \mathbf{R}^{m\times n}$), due to it only needing to minimize the sum of the $\min(m,n)-r$ minimum singular values, where $r$ is the truncated number. 
 
 Utilizing quaternion-based optimization, although the multiplication is uncommunicative in the quaternion domain, color images may be processed as a whole, and, consequently, sparsity may be used simultaneously for image processing in the quaternion domain. In \cite{DBLP:journals/tip/XuYXZN15}, a quaternion-based dictionary-learning algorithm was used on color images sparse representation (SR), while in \cite{DBLP:journals/tip/WangKZT21}, a new robust estimator was used to measure the quaternion residual error for SR. With regard to color image completion, in \cite{DBLP:journals/nla/JiaNS19}, a robust quaternion matrix completion algorithm was provided based on factorizing the target quaternion matrix to the sum of a low-rank quaternion matrix and a sparse quaternion matrix. Although a rigorous analysis was provided for this method, however, the work suggested that it cannot be applied to cases where the missing rate of images is excessively high, and, in addition, the convergence rate of the method is slow. 

In spite of the successes of these algorithms being  encouraging, most of them are optimized only in the real domain; where images are optimized in the quaternion domain, the main direction is SR, and, in terms of image completion, the results as seen in \cite{DBLP:journals/nla/JiaNS19} are not very satisfactory.  Hence, to develop more accurate reconstructions, sparsity as an additional source of information is considered in this paper to facilitate color image completion in the quaternion domain. This sparsity is represented by QDCT , with low-rankness depicted using the quaternion truncated nuclear norm (QTNN) as proposed in our previous work \cite{DBLP:journals/jvcir/YangKM21}. The resulting novel model is named the Low-rank Quaternion Recovery with Sparse Regularization. (LRQR-SR) model. The other main contributions of this work are thus as follows: 
 \begin{itemize}
 	\item{The work focuses on quaternion matrix completion based on combining low-rankness and sparsity. The underlying concept is that the  quaternion-based method can preserve the structure of the color image, allowing the sparsity to be formulated as an $l_1$ norm regularizer in the transformed domain (QDCT).}
 	\item{The closed form solution of  the model, combining the Frobenius norm and the $l_1$ norm, is thus proposed and supported with theoretical analysis.}
 	\item{Extensive experimental results on real color images demonstrate the competitive performance of the proposed method in comparison with several state-of-the-art methods. }
 \end{itemize} 

The outline of the remainder of this article is as follows. Section \ref{P} generalizes some  preliminaries, while Section \ref{Main} offers the proposed LRQR-SR model and the corresponding two-step ADMM-based optimization. Section \ref{E} then presents the experimental results to demonstrate the efficiency of the proposed LRQR-SR, while Section \ref{C} offers a conclusion to the work.

\section{Notations and Preliminaries}
\label{P}
\subsection{Notations}
In the real domain $\mathbb{R}$, we denote scalar, vector, and matrix as a, $\mathbf{a}$, and  $\mathbf{A}$ ,respectively. In the quaternion domain $\mathbb{H}$, we denote scalar, vector, and matrix as $\dot{a}$, $\dot{\mathbf{a}}$, and $\dot{\mathbf{A}}$, respectively. Besides, we denote the complex space as $\mathbb{C}$.   For a quaternion $\dot{q}$, we denote the real part and imaginary part as  $\mathfrak{R}(\dot{q})$ and $\mathfrak{I}(\dot{q})$. We denote transpose, conjugate transpose, and inverse as $\mathbf{(\cdot)}^T$, $\mathbf{(\cdot)}^H$, and $\mathbf{(\cdot)}^{-1}$, respectively. We use $\|\cdot\|_F$ and $\|\cdot\|_*$ to represent the  Frobenius norm  and nuclear norm. We define the inner product of $*_1$ and $*_2$ as $\langle*_1\cdot*_2\rangle \triangleq \text{tr}(*_1^H*_2)$, where $tr(\cdot)$ is the trace function. Both $\textbf{I}_{r\times r}$ and $\textbf{I}_{r}$ denote the $r\times r$ identity matrix.
\subsection{Preliminaries}
Quaternions are proposed by Hamilton in 1843 \cite{doi:10.1080/14786444408644923}. A quaternion number $\dot{q}\in\mathbb{H}$ is combined by a real part and three imagery parts, and can be written as following form:
\begin{equation}
\dot{q}=q_0+q_1\emph{i}+q_2\emph{j}+q_3\emph{k},
\end{equation}
where $q_n\in\mathbb{R}$ $(n=0,1,2,3)$, and \emph{i, j, k} are three imaginary number units which have the following relationships:
\begin{equation*}
\begin{cases}
\emph{i}^2= \emph{j}^2 =\emph{k}^2= \emph{i}\emph{j}\emph{k}=-1\\
\emph{i}\emph{j}=-\emph{j}\emph{i} = \emph{k},    \emph{j}\emph{k}=-\emph{k}\emph{j} = \emph{i},  \emph{k}\emph{i}=-\emph{i}\emph{k} = \emph{j}.
\end{cases}
\end{equation*}
$\mathfrak{R}(\dot{q}) \triangleq q_0$ is the real part of  $\dot{q}$. $\mathfrak{I}(\dot{q}) \triangleq q_1\emph{i}+q_2\emph{j}+q_3\emph{k}$ is the  imaginary part of $\dot{q}$. Hence $\dot{q}=\mathfrak{R}(\dot{q})+\mathfrak{I}(\dot{q})$. Besides, when real part $q_0 = 0$, $\dot{q}$ is  a pure quaternion.  The conjugate and the modulus of $\dot{q}$  are defined as: $\dot{q}^{*} \triangleq q_0-q_1\emph{i}-q_2\emph{j}-q_3\emph{k}$  and $|\dot{q}|\triangleq \sqrt{\dot{q}\dot{q}^{*}}=\sqrt{q_0^2+q_1^2+q_2^2+q_3^2}$. Assuming two quaternions $\dot{p}$ and $\dot{q}\in\mathbb{H}$,  the addition and multiplication are  separately defined as following:
\begin{equation}\nonumber 
\dot{p}+\dot{q}=(p_0+q_0)+(p_1+q_1)\emph{i}+(p_2+q_2)\emph{j}+(p_3+q_3)\emph{k},
\end{equation}
\begin{equation}\nonumber
\begin{aligned}
\dot{p}\dot{q}=& (p_0q_0-p_1q_1-p_2q_2-p_3q_3)\\&+(p_0q_1+p_1q_0+p_2q_3-p_3q_2)\emph{i}
\\&+(p_0q_2-p_1q_3+p_2q_0+p_3q_1)\emph{j}\\&+(p_0q_3+p_1q_2-p_2q_1+p_3q_0)\emph{k}.
\end{aligned}
\end{equation}
It is important to note that the multiplication in the quaternion domain is not commutative  $\dot{p}\dot{q} \neq \dot{p}\dot{q}$.

For quaternion matrix $\dot{\mathbf{Q}}=(\dot{q}_{ij})\in\mathbb{H}^{M \times N}$, where $\dot{\mathbf{Q}}=\mathbf{Q}_0+\mathbf{Q}_1\emph{i}+\mathbf{Q}_2\emph{j}+\mathbf{Q}_3\emph{k}$ and $\mathbf{Q}_n\in\mathbb{R}^{M \times N}$ $(n=0,1,2,3)$ are real matrices. When  $\mathbf{Q}_0 = \mathbf{0}$, $\dot{\mathbf{Q}}$ is a pure quaternion matrix. The Frobenius norm is defined as: $\parallel\dot{\mathbf{Q}}\parallel_F = \sqrt{\sum_{i=1}^{M}\sum_{j=1}^{N}|\dot{q}_{ij}|^2}=\sqrt{tr(\dot{\mathbf{Q}}^H\dot{\mathbf{Q}})}$.

\begin{definition}[\textbf{The Cayley-Dickson form}\label{def1} \cite{DBLP:journals/sigpro/BihanM04}]
The Cayley-Dickson form of quaternion matrix $\dot{\mathbf{Q}}=\mathbf{Q}_0+\mathbf{Q}_1\emph{i}+\mathbf{Q}_2\emph{j}+\mathbf{Q}_3\emph{k}\in\mathbb{H}^{M \times N}$ is $\dot{\mathbf{Q}}=\mathbf{Q}_p+\mathbf{Q}_q\emph{j}$, where $\mathbf{Q}_p=\mathbf{Q}_0+\mathbf{Q}_1\emph{i}$ and $\mathbf{Q}_q=\mathbf{Q}_2+\mathbf{Q}_3\emph{i}\in\mathbb{C}^{M \times N}$. Then the isomorphic complex  matrix representation of  quaternion matrix $\dot{\mathbf{Q}}$ can be denoted as  $\mathbf{Q}_c\in\mathbb{C}^{2M \times 2N}$:
\begin{equation*}\label{ct}
\mathbf{Q}_c={
	\left( \begin{array}{cc}
	\mathbf{Q}_p & \mathbf{Q}_q  \\
	-\mathbf{Q}_q^* & \mathbf{Q}_p^*\\
	\end{array}
	\right )_{2M \times 2N}}.
\end{equation*}
\end{definition}

\begin{definition}[\textbf{The rank of quaternion matrix} \cite{zhang1997quaternions}]The rank of quaternion matrix $\dot{\mathbf{Q}}=(\dot{q}_{ij})\in\mathbb{H}^{M \times N}$ is defined as  the  maximum number of right (left) linearly independent columns (rows)  of  $\dot{\mathbf{Q}}$.
\end{definition}

\begin{theorem}[\textbf{QSVD} \cite{zhang1997quaternions}]
	\label{th1}
	Given a quaternion matrix $\dot{\mathbf{Q}}\in\mathbb{H}^{M \times N}$ be of rank $r$. There are two unitary quaternion matrices $\dot{\mathbf{U}}\in\mathbb{H}^{M \times M}$
	and $\dot{\mathbf{V}}\in\mathbb{H}^{N \times N}$ such that
	\begin{equation*}
	\dot{\mathbf{Q}}=\dot{\mathbf{U}}
	\left( \begin{array}{cc}
	\mathbf{\Sigma}_r & \mathbf{0}  \\
	\mathbf{0} & \mathbf{0}\\
	\end{array}
	\right )\dot{\mathbf{V}}^H= \dot{\mathbf{U}}\mathbf{\Lambda}\dot{\mathbf{V}}^H,
	\end{equation*}
	where $\mathbf{\Sigma}_r=diag({\sigma_1,\cdots, \sigma_r})\in\mathbb{R}^{r\times r}$, and all singular values $\sigma_i>0,  i=1,\cdots,r$. 
\end{theorem}

QSVD and SVD have many properties in common, such as all the singular values are nonnegative, and the order of singular values is decreasing.  As we can see, the rank of the quaternion matrix is the $l_0$ norm of the vector $\{\sigma_i(	\dot{\mathbf{Q}})\}_{i=1}^{\min(M, N)}$, however, the $l_0$ norm is nonconvex so as to QNN is considered, which is similar to the concept of NN in the real domain.  

\begin{definition}[\textbf{QNN} \cite{DBLP:journals/tip/ChenXZ20,DBLP:journals/ijon/YuZY19}]
	\label{def2}
	The nuclear norm of the quaternion matrix  $\dot{\mathbf{Q}}\in\mathbb{H}^{M \times N}$ is defined as $\parallel\dot{\mathbf{Q}}\parallel_*=\sum_{i=1}^{\min(M,N)}\sigma_{i}(\dot{\mathbf{Q}})$, where $\sigma_{i}$ is singular value that can be obtained from the QSVD of  $ \dot{\mathbf{Q}}$.
\end{definition}

As observed in  \cite{DBLP:conf/icip/BihanS03}, the bigger singular values would maintain more information on the color image than smaller singular values. Moreover, the first few largest singular values do not change the rank. Hence, the quaternion-based truncated nuclear norm (QTNN) is developed as follows. 

\begin{definition}[\textbf{QTNN}\cite{DBLP:journals/jvcir/YangKM21}]
	\label{def3}
	The sum of ${\rm min}(M,N)-r$ minimum singular values is the quaternion truncated nuclear norm of the quaternion matrix  $\dot{\mathbf{Q}}\in\mathbb{H}^{M \times N}$, i.e., $\parallel\dot{\mathbf{Q}}\parallel_r=\sum_{i=r+1}^{{\rm min}(M,N)}\sigma_{i}(\dot{\mathbf{Q}})$.
\end{definition}

\begin{theorem}[\cite{DBLP:journals/jvcir/YangKM21}]
	\label{th2}
	For any quaternion matrix $\dot{\mathbf{Q}}\in\mathbb{H}^{M \times N}$, and any matrices $\dot{\mathbf{A}}\in\mathbb{H}^{r \times M}$  and $\dot{\mathbf{B}}\in\mathbb{H}^{r \times N}$ that are satisfied with $\dot{\mathbf{A}}\dot{\mathbf{A}}^{H}=\mathbf{I}_{r\times r}$ and $\dot{\mathbf{B}}\dot{\mathbf{B}}^{H}=\mathbf{I}_{r\times r}$. $r$ is any positive integer $(r\leq {\rm min}(M,N))$, we have
	\begin{equation}
	\mid tr(\dot{\mathbf{A}}\dot{\mathbf{Q}}\dot{\mathbf{B}}^{H})\mid\leq\sum_{i=1}^r\sigma_{i}(\dot{\mathbf{Q}}).
	\end{equation}
	Besides, 
	$
	{\rm max} |tr(\dot{\mathbf{A}}\dot{\mathbf{Q}}\dot{\mathbf{B}}^H)|=\sum_{i=1}^r\sigma_{i}(\dot{\mathbf{Q}}).
	$
\end{theorem}

\section{Low-rank Quaternion Recovery with Sparse Regularization}
\label{Main}
This section is divided into three parts. Subsection \ref{3.1} gives the formulation of the proposed LRQR-SR, while subsection \ref{3.2} presents the quaternion discrete cosine transform that we used in this paper, and subsection \ref{3.3} gives the corresponding ADMM-based optimization algorithm.
\subsection{Problem Formulation}
\label{3.1}
This section introduces the formulation of a low-rank quaternion optimized model for color image recovery. 

Let $\dot{\mathbf{O}}\in\mathbb{H}^{M \times N}$ be the partial observed color image. The purpose of the resulting problem is thus to recover $\dot{\mathbf{X}}\in \mathbb{H}^{M\times N}$ by improving QNN as applied to the characterization of the low-rank property more accurately. The QNN model can be formulated as
\begin{equation}\label{m1}
\begin{aligned}
&\min\limits_{\dot{\mathbf{X}}} \parallel\dot{\mathbf{X}}\parallel_*\\& \text{ s.t.}   \quad   P_\Omega(\dot{\mathbf{X}})=P_\Omega(\dot{\mathbf{O}}),
\end{aligned}
\end{equation}
where $\Omega$ is the index set of the observed data, and the linear operation $P_\Omega(*)$ is the operator that indicates that the elements in $\Omega$ are remain while other elements are zero. 

As the several largest singular values will not influence the rank, the TQNN model is proposed, and the estimation of  low-rankness should thus be more accurate.  Based on model \eqref{m1} and Definition \ref{def3}, the QTNN model can thus be formulated as
\begin{equation}\label{m2}
\begin{aligned}
&\min\limits_{\dot{\mathbf{X}}}\parallel\dot{\mathbf{X}}\parallel_r   \\&  \text{s.t.}   \quad   P_\Omega(\dot{\mathbf{X}}-\dot{\mathbf{O}})=0,
\end{aligned}
\end{equation} 
where $\|\dot{\mathbf{X}}\|_r=\sum_{i=r+1}^{\min(m,n)}\sigma_{i}(\dot{\mathbf{X}})=\sum_{i=1}^{\min(m,n)}\sigma_{i}(\dot{\mathbf{X}})-\sum_{i=r}^{\min(m,n)}\sigma_{i}(\dot{\mathbf{X}})=\parallel\dot{\mathbf{X}}\parallel_*-\sum_{i=r}^{\min(m,n)}\sigma_{i}(\dot{\mathbf{X}})$. 

Being low rank is a necessary but insufficient condition for MC \cite{DBLP:conf/eccv/LiangRZM12}.  Motivated by this observation and the success of sparsity as utilized in  MC  \cite{DBLP:journals/spic/DongXGHW18, DBLP:conf/eccv/XuZZ18, DBLP:journals/pr/FanC17a} the LRQR-SR model is developed, which considers both a  low-rank constraint and sparsity in the quaternion domain. The low-rank constraint is depicted using QTNN in this case, while  sparsity is depicted using the $l_1$ norm. As in the previously mentioned strategies, the quaternion matrix is assumed to be sparse in a certain transformed domain.  Hence, the resulting model can be formulated as 
 \begin{equation}\label{m3}
 \begin{aligned}
 &\min\limits_{\dot{\mathbf{X}}}\parallel\dot{\mathbf{X}}\parallel_*-\sum_{i=r}^{\min(m,n)}\sigma_{i}(\dot{\mathbf{X}})+\lambda \parallel\dot{\mathbf{D}}\parallel_1\\&\text{s.t.}  \quad P_\Omega(\dot{\mathbf{X}}-\dot{\mathbf{O}})=0 \\&\qquad \mathcal{T}(\dot{\mathbf{X}})=\dot{\mathbf{D}}
 \end{aligned}
 \end{equation} 
 where $\mathcal{T}(\cdot)$ is the transform operator, $\dot{\mathbf{D}}$ is the transformed quaternion matrix, and $\lambda$ is a positive number. 
 
 However, it is hard to solve problem \eqref{m3} directly due to the fact that  QTNN is nonconvex. To address this problem, Theorem \ref{th2} is applied such that problem \eqref{m3} can be rewritten as 
  \begin{equation}\label{m4}
 \begin{aligned}
 &\min\limits_{\dot{\mathbf{X}}}\parallel\dot{\mathbf{X}}\parallel_*- \mathop{\max}\limits_{\dot{\mathbf{A}}\dot{\mathbf{A}}^H=\mathbf{I},\dot{\mathbf{B}}\dot{\mathbf{B}}^H=\mathbf{I}} |tr(\dot{\mathbf{A}}\dot{\mathbf{X}}\dot{\mathbf{B}}^H)| +\lambda \parallel\dot{\mathbf{D}}\parallel_1\\&\text{s.t.}  \quad P_\Omega(\dot{\mathbf{X}}-\dot{\mathbf{O}})=0 \\&\qquad \mathcal{T}(\dot{\mathbf{X}})=\dot{\mathbf{D}},
 \end{aligned}
 \end{equation} 
 where $\dot{\mathbf{A}}=(\dot{\mathbf{u}}_{1},\cdots \dot{\mathbf{u}}_{r})^H$ and $\dot{\mathbf{B}}=(\dot{\mathbf{v}}_{1},\cdots \dot{\mathbf{v}}_{r})^H$. $\{\dot{\mathbf{u}}_{1},\cdots\dot{\mathbf{u}}_{r}\}$ and $ \{\dot{\mathbf{v}}_{1},\cdots\dot{\mathbf{v}}_{r}\}$ are the first r columns of  $\dot{\mathbf{U}}$ and $\dot{\mathbf{V}}$. $\dot{\mathbf{U}}$ and $\dot{\mathbf{V}}$ are left and right unitary quaternion matrices that are calculated by QSVD of  $\dot{\mathbf{X}}$.  
 
 In this way, the whole procedure of the method can be divided into two main steps: in the first step, the quaternion matrices are computed by QSVD, and then the main goal becomes to optimize problem \eqref{m4}. The overall procedure is summarized in Algorithm \ref{a1}. 
 \begin{algorithm}[htbp]
 	\caption{Low-rank Quaternion Recovery with Sparse Regularization}
 	\label{a1}
 	\begin{algorithmic}[1]
 		\REQUIRE   the observed quaternion matrix  $\dot{\mathbf{O}}\in\mathbb{H}^{M\times N}$, the position set of observed elements $\Omega$, and the tolerance $\varepsilon_0$.
 		\STATE \textbf{Initial} the initial number of iteration $k=1$, $\dot{\mathbf{X}}_1=\dot{\mathbf{O}}$.
 		\STATE \textbf{Repeat}
 		\STATE \quad \textbf{Step 1.} Calculating the QSVD of the given $\dot{\mathbf{X}}_{k}$\\
 		\qquad  \qquad \qquad  $[\dot{\mathbf{U}}_{k},\mathbf{\Sigma}_{k},\dot{\mathbf{V}}_{k}]=\text{QSVD}(\dot{\mathbf{X}}_{k})$
 		\STATE  \quad where $\dot{\mathbf{U}}_{k}=(\dot{\mathbf{u}}_{1},\cdots, \dot{\mathbf{u}}_{m})\in\mathbb{H}^{M\times M}$,\\
 		\qquad \qquad $\dot{\mathbf{V}}_{k}=(\dot{\mathbf{v}}_{1},\cdots, \dot{\mathbf{v}}_{n})\in\mathbb{H}^{N\times N}$.
 		\STATE   \quad Calculating
 	     $\dot{\mathbf{A}}_k=(\dot{\mathbf{u}}_{1},\cdots, \dot{\mathbf{u}}_{r})^T\in\mathbb{H}^{r\times M}$ and \\ 
 	     \qquad\qquad\qquad$\dot{\mathbf{B}}_k=(\dot{\mathbf{v}}_{1},\cdots, \dot{\mathbf{v}}_{r})^T\in\mathbb{H}^{r\times N}$.
 		
 		\STATE \quad \textbf{Step 2.} Solving the optimization problem as followed \\
 	\quad $\dot{\mathbf{X}}_{k+1}=\arg\min\limits_{\dot{\mathbf{X}}}\parallel\dot{\mathbf{X}}\parallel_*-|tr(\dot{\mathbf{A}}_k\dot{\mathbf{X}}\dot{\mathbf{B}}_k^H)| +\lambda \parallel\dot{\mathbf{D}}\parallel_1$,\\
 		\qquad$\text{s.t.}  \quad P_\Omega(\dot{\mathbf{X}}-\dot{\mathbf{O}})=0 \qquad \mathcal{T}(\dot{\mathbf{X}})=\dot{\mathbf{D}}.$
 	 
 		\STATE \textbf{Until convergence} $\|\dot{\mathbf{X}}_{k+1}-\dot{\mathbf{X}}_{k}\|_F \leq \varepsilon_0$, $\dot{\mathbf{X}}_{opt}=\dot{\mathbf{X}}_{k+1}$.
 		\ENSURE  the recovered quaternion matrix $\dot{\mathbf{X}}_{opt}$.
 	\end{algorithmic}
 \end{algorithm}

\subsection{Quaternion Discrete Cosine Transform}
\label{3.2}
Several key points of utilizing QDCT are outlined below.
\subsubsection{The reasons for utilizing QDCT }
Most importantly, the proposed method operates in the quaternion domain, where each color image can be handled by quaternion algebra as a whole. In order to avoid destroying the RGB structure and to improve the accuracy of image recovery, the entire process must thus be operated in the quaternion domain. In addition, the spectral coefficients obtained using QDCT have strong energy and good redundancy elimination characteristics \cite{DBLP:journals/ijon/LiMXLZ17}, while QDCT itself is easy to quantitatively analyze. Finally, in the real and complex fields, the energy concentration of the input information throughout two-dimensional DCT is higher than that of the input information after DFT. Research into QDCT is driven by the existence of successful applications in both the real and complex domains, and for these reasons, QDCT is adopted in the proposed method. 

\begin{figure} [htbp]
 	\centering
 	\includegraphics[width=90mm]{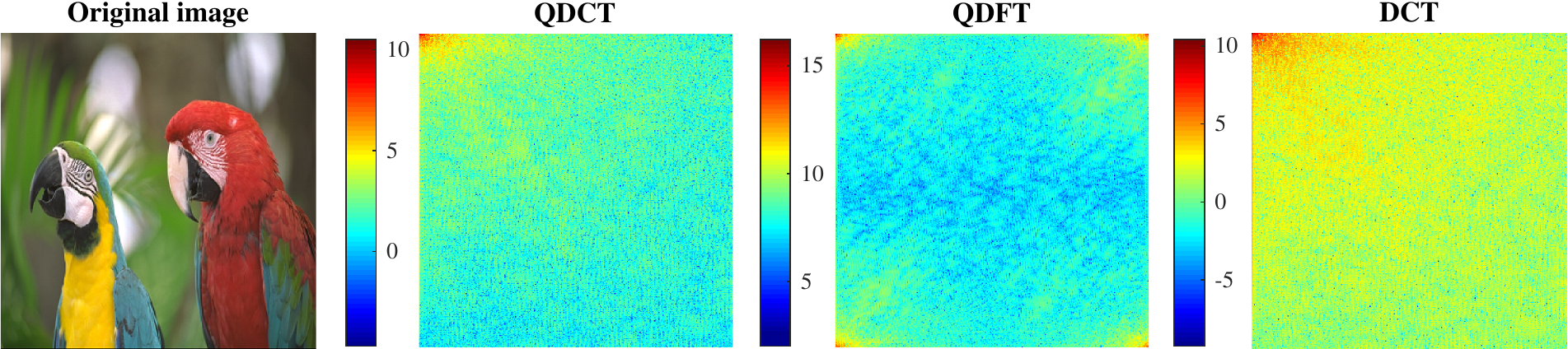}
 	\caption{The comparison of energy compaction  after different transformation}
 	\label{qdct}
 \end{figure}

Fig. \ref{qdct} gives an illustration of the proposed sparse representation on the color image “$\mathit{Parrot}$”. The first image is the original image; the second, third, and fourth image respectively display the coefficients after QDCT, QDFT, and DCT (grayscale image) using a logarithmic scale. After transformation, the coefficients of QDCT and DCT are mainly concentrated in the upper left corner, and most of the remaining coefficients are close to zero. However, after QDFT transformation, the coefficients are mainly concentrated at four corners, which means that utilizing cosine transform is superior to Fourier transform to process images in the quaternion domain. Besides, when comparing QDCT with DCT, especially in the upper left corner, it can be observed that QDCT has higher energy compaction than DCT.
\subsubsection{Definition of QDCT} As the multiplication of quaternions is non-commutative, there are two forms of QDCT:  
a left-handed form $\text{QDCT}_L$ and a right-handed form $\text{QDCT}_R$2). These can be formulated as the following equations, respectively \cite{feng2008quaternion}:
\begin{equation}
\text{QDCT}_L(p,s)=\alpha(p)\alpha(s)\sum_{m=0}^{M-1}\sum_{n=0}^{N-1}\dot{u}\cdot\dot{\mathbf{F}}(m,n)\cdot C(p,s,m,n)
\end{equation}
\begin{equation}
  \text{QDCT}_R(p,s)=\alpha(p)\alpha(s)\sum_{m=0}^{M-1}\sum_{n=0}^{N-1}\dot{\mathbf{F}}(m,n)\cdot C(p,s,m,n)\cdot \dot{u},
   \end{equation}
 where $\dot{\mathbf{F}}(m,n)\in\mathbb{H}^{M\times N}$, m and n is the row and column of quaternion matrix $\dot{\mathbf{F}}$. $\dot{u}$ is a pure quaternion and satisfies $\dot{u}^2=-1$. The values of of $\alpha(p), \alpha(s)$ and $C(p,s,m,n)$ are analogous to DCT in the real domain:
\begin{equation*}
\alpha(p)= \left\{
\begin{array}{l}
\sqrt{1/M}\quad p=0 \\  \sqrt{2/M}\quad p\neq0,
\end{array}
\right.
  \alpha(s)=\left\{
\begin{array}{l}
\sqrt{1/N}\quad s=0 \\  \sqrt{2/N}\quad s\neq0
\end{array}
\right.
\end{equation*}
\begin{equation*}
C(p,s,m,n)=\cos[\frac{\pi(2m+1)p}{2M}]\cos[\frac{\pi(2n+1)p}{2N}].
\end{equation*}

Besides, the corresponding inverse transformation of QDCT is the Inverse Quaternion Discrete Cosine Transform (IQDCT). These are thus the transformation pairs of each other, and satisfy the following relationship:
\begin{equation*}
\dot{\mathbf{F}}(m,n)=\text{IQDCT}_L[\text{QDCT}_L(\dot{\mathbf{F}}(m,n))]
\end{equation*}
\begin{equation*}
\dot{\mathbf{F}}(m,n)=\text{IQDCT}_R[\text{QDCT}_R(\dot{\mathbf{F}}(m,n))].
\end{equation*}

In the proposed algorithm, $\text{QDCT}_L$ is utilized to calculate QDCT.  

\subsubsection{Calculation of  $\text{QDCT}_L$}
To simplify the calculation of QDCT, we take full advantage of the Cayley Dickson form seen in Definition \ref{def1}, as in \cite{feng2008quaternion} is used. The whole process of $\text{QDCT}_L$ calculation is as follows:
\begin{enumerate}
	\item[a] Transforming the given quaternion matrix $\dot{\mathbf{F}}(m,n)\in\mathbb{H}^{M\times N}$ to the Cayley Dickson form $\dot{\mathbf{F}}(m,n)=\mathbf{F}_p(m,n)+\mathbf{F}_q(m,n)\emph{j}$, where $\mathbf{F}_p(m,n)$ and $\mathbf{F}_q(m,n)\in\mathbb{C}^{M \times N}$ 
	\item[b] Calculating the DCT of complex matrices $\mathbf{F}_p(m,n)$ and $\mathbf{F}_q(m,n)$. The results are denoted as $\text{DCT}_C(\mathbf{F}_p(m,n))$ and $\text{DCT}_C(\mathbf{F}_q(m,n))$, respectively.
	\item[c] Using $\text{DCT}_C(\mathbf{F}_p(m,n))$ and $\text{DCT}_C(\mathbf{F}_q(m,n))$ to form a quaternion matrix: $\dot{\mathbf{F}}^{'}(m,n)=\text{DCT}_C(\mathbf{F}_p(m,n))+\text{DCT}_C(\mathbf{F}_q(m,n))j$.
	\item[d] Multiplying $\dot{\mathbf{F}}^{'}(m,n)$ with the quaternion factor $\dot{u}$ to get the final result $\text{QDCT}_L$:\\
	$\text{QDCT}_L(\dot{\mathbf{F}}(m,n))=\dot{u}\cdot\dot{\mathbf{F}}^{'}(m,n)$.
\end{enumerate}

\subsection{ADMM-Based Optimization Algorithm}
\label{3.3}
Following the model we discussed in subsection \ref{3.1} and the transformation introduced in \ref{3.2}, ADMM was adopted to optimize problem \eqref{m4}. This involves introducing auxiliary variable $\dot{\mathbf{H}}$  and reformulating \eqref{m4} as 
\begin{equation}\label{m5}
\begin{aligned}
&\min\limits_{\dot{\mathbf{X}}}\parallel\dot{\mathbf{X}}\parallel_*- \mathop{\max}\limits_{\dot{\mathbf{A}}\dot{\mathbf{A}}^H=\mathbf{I},\dot{\mathbf{B}}\dot{\mathbf{B}}^H=\mathbf{I}} |tr(\dot{\mathbf{A}}\dot{\mathbf{H}}\dot{\mathbf{B}}^H)| +\lambda \parallel\dot{\mathbf{D}}\parallel_1\\&\text{s.t.}  \quad P_\Omega(\dot{\mathbf{X}}-\dot{\mathbf{O}})=0 \\&\qquad \dot{\mathbf{H}}=\dot{\mathbf{X}}
\\&\qquad \mathcal{T}_{QDCT_L}(\dot{\mathbf{X}})=\dot{\mathbf{D}}.
\end{aligned}
\end{equation} 
In analogy with the ADMM framework adopted in the complex domain \cite{li2015alternating},as the multiplication is not commutative in the quaternion domain, the augmented Lagrangian function of  \eqref{m5} can be written as
\begin{equation}\label{m6}
\begin{split}
&L(\dot{\mathbf{X}},\dot{\mathbf{H}},\dot{\mathbf{D}},\dot{\mathbf{Y}},\dot{\mathbf{Z}}, \beta)=\\&\parallel\dot{\mathbf{X}}\parallel_*- |tr(\dot{\mathbf{A}}\dot{\mathbf{H}}\dot{\mathbf{B}}^H)|+\lambda \parallel\dot{\mathbf{D}}\parallel_1\\&+\mathfrak{R}(tr(\dot{\mathbf{Y}}^H(\dot{\mathbf{X}}-\dot{\mathbf{H}})))+\frac{\beta}{2}\parallel\dot{\mathbf{X}}-\dot{\mathbf{H}}\parallel_F^2\\&+\mathfrak{R}(tr(\dot{\mathbf{Z}}^H(\dot{\mathbf{D}}-\mathcal{T}_{QDCT_L}(\dot{\mathbf{X}}))))\\&
+\frac{\beta}{2}\parallel\dot{\mathbf{D}}-\mathcal{T}_{QDCT_L}(\dot{\mathbf{X}})\parallel_F^2,
\end{split}
\end{equation}
where $\dot{\mathbf{Y}}$ and  $\dot{\mathbf{Z}}$ are the Lagrange multipliers, and $\beta$ is the positive penalty parameter. Under the framework of ADMM, the variables $\dot{\mathbf{X}}$, $\dot{\mathbf{H}}$, $\dot{\mathbf{D}}$, $\dot{\mathbf{Y}}$, and $\dot{\mathbf{Z}}$ are updated alternately in the $p$-th iteration using		
\begin{equation}
\left\{
\begin{array}{lr}
\dot{\mathbf{X}}^{p+1}=\arg\min\limits_{\dot{\mathbf{X}}}L(\dot{\mathbf{X}},\dot{\mathbf{H}}^{p},\dot{\mathbf{D}}^{p},\dot{\mathbf{Y}}^{p},\dot{\mathbf{Z}}^{p}, \beta^p), &  \\
\dot{\mathbf{D}}^{p+1}=\arg\min\limits_{\dot{\mathbf{D}}}L(\dot{\mathbf{X}}^{p+1},\dot{\mathbf{H}}^{p},\dot{\mathbf{D}},\dot{\mathbf{Y}}^{p},\dot{\mathbf{Z}}^{p}, \beta^p),\\
\dot{\mathbf{H}}^{p+1}=\arg\min\limits_{\dot{\mathbf{H}}}L(\dot{\mathbf{X}}^{p+1},\dot{\mathbf{H}}, \dot{\mathbf{D}}^{p+1},\dot{\mathbf{Y}}^{p},\dot{\mathbf{Z}}^{p}, \beta^p),\\
\dot{\mathbf{Y}}^{p+1}=\dot{\mathbf{Y}}^{p}+\beta^p(\dot{\mathbf{X}}^{p+1}-\dot{\mathbf{H}}^{p+1}),\\
\dot{\mathbf{Z}}^{p+1}=\dot{\mathbf{Z}}^{p}+\beta^p(\dot{\mathbf{D}}^{p+1}-\mathcal{T}_{QDCT_L}(\dot{\mathbf{X}}^{p+1})).
\end{array}
\right.
\end{equation}

The $\dot{\mathbf{X}}$ subproblem is
\begin{equation}\label{m7}
\begin{split}
\dot{\mathbf{X}}^{p+1}&=\arg\min\limits_{\dot{\mathbf{X}}}\parallel\dot{\mathbf{X}}\parallel_*+\mathfrak{R}(tr(\dot{\mathbf{Y}}^{pH}(\dot{\mathbf{X}}-\dot{\mathbf{H}}^p)))\\&\quad+\frac{\beta^p}{2}\parallel\dot{\mathbf{X}}-\dot{\mathbf{H}}^p\parallel_F^2+\mathfrak{R}(tr(\dot{\mathbf{Z}}^{pH}(\dot{\mathbf{D}}^p-\mathcal{T}_{QDCT_L}(\dot{\mathbf{X}}))))\\&\quad
+\frac{\beta^p}{2}\parallel\dot{\mathbf{D}}^p-\mathcal{T}_{QDCT_L}(\dot{\mathbf{X}})\parallel_F^2
\\&=\arg\min\limits_{\dot{\mathbf{X}}}\parallel\dot{\mathbf{X}}\parallel_*+\frac{\beta^p}{2}\parallel\dot{\mathbf{X}}-\dot{\mathbf{H}}^p+\dot{\mathbf{Y}}^p/\beta^p\parallel_F^2\\&\quad+\frac{\beta^p}{2}\parallel\dot{\mathbf{D}}^p-\mathcal{T}_{QDCT_L}(\dot{\mathbf{X}})+\dot{\mathbf{Z}}^p/\beta^p\parallel_F^2.
\end{split}
\end{equation}
In the last term of  \eqref{m7}, $\dot{\mathbf{X}}$ can not be separated directly as the transformation. Despite this, the Parseval theorem in the quaternion domain indicates that the total energy of signal computed in the
quaternionic domain and total energy of signal computed in the spatial domain must be the same \cite{DBLP:journals/cma/BahriHHA08, DBLP:journals/corr/Hitzer13d}. This means that a unitary transformation preserves energy conservation under the Frobenius norm, and thus the last term of \eqref{m7} can be rewritten as
\begin{equation*}
\begin{split}
&\frac{\beta^p}{2}\parallel\dot{\mathbf{D}}^p-\mathcal{T}_{QDCT_L}(\dot{\mathbf{X}})+\dot{\mathbf{Z}}^p/\beta^p\parallel_F^2\\&=\frac{\beta^p}{2}\parallel\mathcal{T}_{IQDCT_L}(\dot{\mathbf{D}}^p+\dot{\mathbf{Z}}^p/\beta^p)-\dot{\mathbf{X}}\parallel_F^2,
\end{split}
\end{equation*}
where the $\mathcal{T}_{IQDCT_L}$ is the inverse transformation of $\text{QDCT}_L$. For a more concise representation, let $\mathcal{T}$ denote $\mathcal{T}_{QDCT_L}$ and $\mathcal{IT}$ denote $\mathcal{T}_{IQDCT_L}$. Consequently, \eqref{m7} can be reformulated as
\begin{equation*}
\begin{split}
\dot{\mathbf{X}}^{p+1}&=\arg\min\limits_{\dot{\mathbf{X}}}\parallel\dot{\mathbf{X}}\parallel_*+\frac{\beta_k}{2}\parallel\dot{\mathbf{X}}-\dot{\mathbf{H}}^p+\dot{\mathbf{Y}}^p/\beta^p\parallel_F^2\\&\quad+\frac{\beta^p}{2}\parallel\mathcal{IT}(\dot{\mathbf{D}}^p+\dot{\mathbf{Z}}^p/\beta^p)-\dot{\mathbf{X}}\parallel_F^2\\&=\arg\min\limits_{\dot{\mathbf{X}}}\parallel\dot{\mathbf{X}}\parallel_*\\&\quad+\beta^p\parallel\dot{\mathbf{X}}-\frac{1}{2}[\dot{\mathbf{H}}^p+\dot{\mathbf{Y}}^p/\beta^p+\mathcal{IT}(\dot{\mathbf{D}}^p+\dot{\mathbf{Z}}^p/\beta^p)]\parallel_F^2.
\end{split}
\end{equation*}
The closed solution of the above problem is
\begin{equation}\label{m8}
\dot{\mathbf{X}}^{p+1}= \mathfrak{D}_{\frac{1}{2\beta^p}}(\frac{1}{2}[\dot{\mathbf{H}}^p+\dot{\mathbf{Y}}^p/\beta^p+\mathcal{IT}(\dot{\mathbf{D}}^p+\dot{\mathbf{Z}}^p/\beta^p)]),
\end{equation}
where $\mathfrak{D}_{\tau}(*) $ is the quaternion singular value shrinkage operator \cite{DBLP:journals/tip/ChenXZ20} is defined as 
\begin{equation*}
\mathfrak{D}_{\tau}(\dot{\mathbf{A}})=\dot{\mathbf{U}}\mathfrak{D}_{\tau}(\mathbf{\Sigma})\dot{\mathbf{V}}^H, \mathfrak{D}_{\tau}(\mathbf{\Sigma})=diag({\rm max}\{\sigma_i-\tau,0\}),
\end{equation*}
where $\dot{\mathbf{U}}$ ,  $\dot{\mathbf{V}}$, and $\sigma_i$ are obtained by computing QSVD of quaternion matrix $
\dot{\mathbf{A}}=\dot{\mathbf{U}}\mathbf{\Sigma}\dot{\mathbf{V}}^H, \mathbf{\Sigma}=diag({\sigma_1,\cdots, \sigma_r},0\cdots, 0 )\in\mathbb{R}^{M\times N}.$

The $\dot{\mathbf{D}}$ subproblem is 
\begin{equation}\label{m9}
\begin{split}
\dot{\mathbf{D}}^{p+1}&=\arg\min\limits_{\dot{\mathbf{D}}}\lambda \parallel\dot{\mathbf{D}}\parallel_1+\mathfrak{R}(tr(\dot{\mathbf{Z}}^{pH}(\dot{\mathbf{D}}-\mathcal{T}(\dot{\mathbf{X}}^{p+1}))))\\&\quad
+\frac{\beta^p}{2}\parallel\dot{\mathbf{D}}-\mathcal{T}(\dot{\mathbf{X}}^{p+1})\parallel_F^2\\&
=\arg\min\limits_{\dot{\mathbf{D}}}\lambda \parallel\dot{\mathbf{D}}\parallel_1+\frac{\beta^p}{2}\parallel\dot{\mathbf{D}}-\mathcal{T}(\dot{\mathbf{X}}^{p+1})+\dot{\mathbf{Z}}^{pH}/\beta^p\parallel_F^2
\end{split}
\end{equation}

To obtain the optimal solution of \eqref{m9}, we have the following theorem. 

\begin{theorem}
	\label{th3}
	For any $\lambda>0$, the closed solution of problem $\min\limits_{\dot{\mathbf{X}}}\lambda \parallel\dot{\mathbf{X}}\parallel_1+\parallel\dot{\mathbf{Y}}-\dot{\mathbf{X}}\parallel_F^2$ can be given by
	\begin{equation}\label{eqth3}
	\dot{\mathbf{X}}_{opt}=\mathcal{S}_{2\lambda}(\dot{\mathbf{Y}}),
	\end{equation}
	where $\mathcal{S}_{\tau}(\cdot)$ represents the element-wise soft thresholding operator defined by
	\begin{equation}
	\mathcal{S}_{\tau}(\dot{\mathbf{x}})=\frac{\dot{\mathbf{x}}}{\mid\dot{\mathbf{x}}\mid}\max\{\mid\dot{\mathbf{x}}\mid-\tau,0\}.
	\end{equation}
\end{theorem}

The proof of Theorem \ref{th3} is given in the Appendix. Based on Theorem \ref{th3}, problem \eqref{m9} has a closed-form solution given by
\begin{equation}\label{m10}
\dot{\mathbf{D}}^{p+1}=\mathcal{S}_{\frac{4\lambda}{\beta^p}}(\mathcal{T}(\dot{\mathbf{X}}^{p+1})-\dot{\mathbf{Z}}^{pH}/\beta^p).
\end{equation}

The $\dot{\mathbf{H}}$ subproblem is 
\begin{equation}\label{m11}
\begin{split}
\dot{\mathbf{H}}^{p+1}&=\arg\min\limits_{\dot{\mathbf{H}}}- |tr(\dot{\mathbf{A}}\dot{\mathbf{H}}\dot{\mathbf{B}}^H)|+\mathfrak{R}(tr(\dot{\mathbf{Y}}^{pH}(\dot{\mathbf{X}}^{p+1}-\dot{\mathbf{H}})))\\&\quad+\frac{\beta}{2}\parallel\dot{\mathbf{X}}^{p+1}-\dot{\mathbf{H}}\parallel_F^2\\&
=\arg\min\limits_{\dot{\mathbf{H}}}\frac{\beta}{2}\parallel\dot{\mathbf{X}}^{p+1}-\dot{\mathbf{H}}+\dot{\mathbf{Y}}^p/\beta^p+\dot{\mathbf{A}}^H\dot{\mathbf{B}}/\beta^p\parallel_F^2.
\end{split}
\end{equation}
Following the above equation, we can obtain 
\begin{equation}
\dot{\mathbf{H}}^{p+1}=\dot{\mathbf{X}}^{p+1}+\dot{\mathbf{Y}}^p/\beta^p+\dot{\mathbf{A}}^H\dot{\mathbf{B}}/\beta^p.
\end{equation} 
Moreover, the observed data should remain unchanged in each iteration such that
\begin{equation}
\dot{\mathbf{H}}^{p+1}=P_{\Omega^C}(\dot{\mathbf{H}}^{p+1})+P_\Omega(\dot{\mathbf{O}}).
\end{equation} 

The update of penalty parameter $\beta^p$ is
\begin{equation}
\beta^{p+1}=\min\{\rho\beta^{p}, \beta_{\max}\},
\end{equation}
where $\beta_{\max}$ is the given maximum value of the penalty parameter, and $\rho \geq1$ is a constant parameter.

\eqref{m5} is the Step 2 problem listed in Algorithm \ref{a1}, so that the whole procedure to solve it is summarized in Algorithm \ref{a2}. 
\begin{algorithm}[htbp]
	\caption{ADMM solver for problem \eqref{m5} in Step 2. }
	\label{a2}
	\begin{algorithmic}[1]
	\REQUIRE    $\dot{\mathbf{O}}$, $\Omega$, $\dot{\mathbf{A}}_l$, $\dot{\mathbf{B}}_l$, tolerance $\varepsilon$, and parameters $\lambda$, $\rho$, $\beta_{\max}$.
	\STATE \textbf{Initial} $\dot{\mathbf{X}}^1=\dot{\mathbf{O}}$, $\dot{\mathbf{H}}^1=\dot{\mathbf{D}}^1=\dot{\mathbf{X}}^1$, and $\beta^1$. Let $\dot{\mathbf{Y}}^1$ and $\dot{\mathbf{Z}}^1$ be random quaternion matrix with the same size of $\dot{\mathbf{X}}^1$.
	\STATE \textbf{Repeat}
	\STATE Update $\dot{\mathbf{X}}^{p+1}$\\ $\dot{\mathbf{X}}^{p+1}=\mathfrak{D}_{\frac{1}{2\beta^p}}(\frac{1}{2}[\dot{\mathbf{H}}^p+\dot{\mathbf{Y}}^p/\beta^p+\mathcal{IT}(\dot{\mathbf{D}}^p+\dot{\mathbf{Z}}^p/\beta^p)]).$
	\STATE  Update $\dot{\mathbf{D}}^{p+1}=\mathcal{S}_{\frac{4\lambda}{\beta^p}}(\mathcal{T}(\dot{\mathbf{X}}^{p+1})-\dot{\mathbf{Z}}^{pH}/\beta^p).$
	\STATE   Update $\dot{\mathbf{H}}^{p+1}=\dot{\mathbf{X}}^{p+1}+\dot{\mathbf{Y}}^p/\beta^p+\dot{\mathbf{A}}^H\dot{\mathbf{B}}/\beta^p,$\\
	\qquad\qquad$\dot{\mathbf{H}}^{p+1}=P_{\Omega^C}(\dot{\mathbf{H}}^{p+1})+P_\Omega(\dot{\mathbf{O}}).$
	\STATE Update $\dot{\mathbf{Y}}^{p+1}=\dot{\mathbf{Y}}^{p}+\beta^p(\dot{\mathbf{X}}^{p+1}-\dot{\mathbf{H}}^{p+1}).$
	\STATE Update
	$\dot{\mathbf{Z}}^{p+1}=\dot{\mathbf{Z}}^{p}+\beta^p(\dot{\mathbf{D}}^{p+1}-\mathcal{T}(\dot{\mathbf{X}}^{p+1})).$
	\STATE Update  $\beta^{p+1}=\min\{\rho\beta^{p}, \beta_{\max}\},$.
	\STATE \textbf{Until convergence} $\|\dot{\mathbf{X}}^{p+1}-\dot{\mathbf{X}}^{p}\|_F \leq \varepsilon$ or $p$ reaches the set maximum iteration  number. 
	\ENSURE  $\dot{\mathbf{X}}^{p+1}$ 
\end{algorithmic}
\end{algorithm}
\section{Experimental Results}
\label{E}
In this section, the effectiveness of the proposed LRQR-SR method is demonstrated in comparison with various relevant state-of-the-art methods  is shown. Subsection \ref{es} provides the experimental settings. Subsection \ref{clr} presents the color image recovery results. Finally, the experimental results are discussed in Subsection \ref{d}. 
\subsection{Experimental Settings} \label{es}
\subsubsection{Comparison Methods} Several relevant existing algorithms were used as comparison algorithms, including D-N and F-N \cite{DBLP:journals/pami/ShangCLLL18},  TNNR \cite{DBLP:journals/pami/HuZYLH13}, TNN-SR \cite{DBLP:journals/spic/DongXGHW18}, Q-DNN  and Q-FNN \cite{DBLP:journals/tsp/MiaoK20}, LRQA \cite{DBLP:journals/tip/ChenXZ20}, QTNN \cite{DBLP:journals/jvcir/YangKM21}. The first four of these algorithms are matrix-based, while the last four algorithms are quaternion-based. D-N, F-N, Q-DNN, and Q-FNN use factorization to depict low-rankness, while LRQA is based on the developed QNN to depict low-rankness, and TNNR, TNN-SR, and QTNN utilize a truncated nuclear norm.
\subsubsection{Test Data and Experimental Environment}
\begin{figure}[htbp]
	\centering
	\includegraphics[width=80mm]{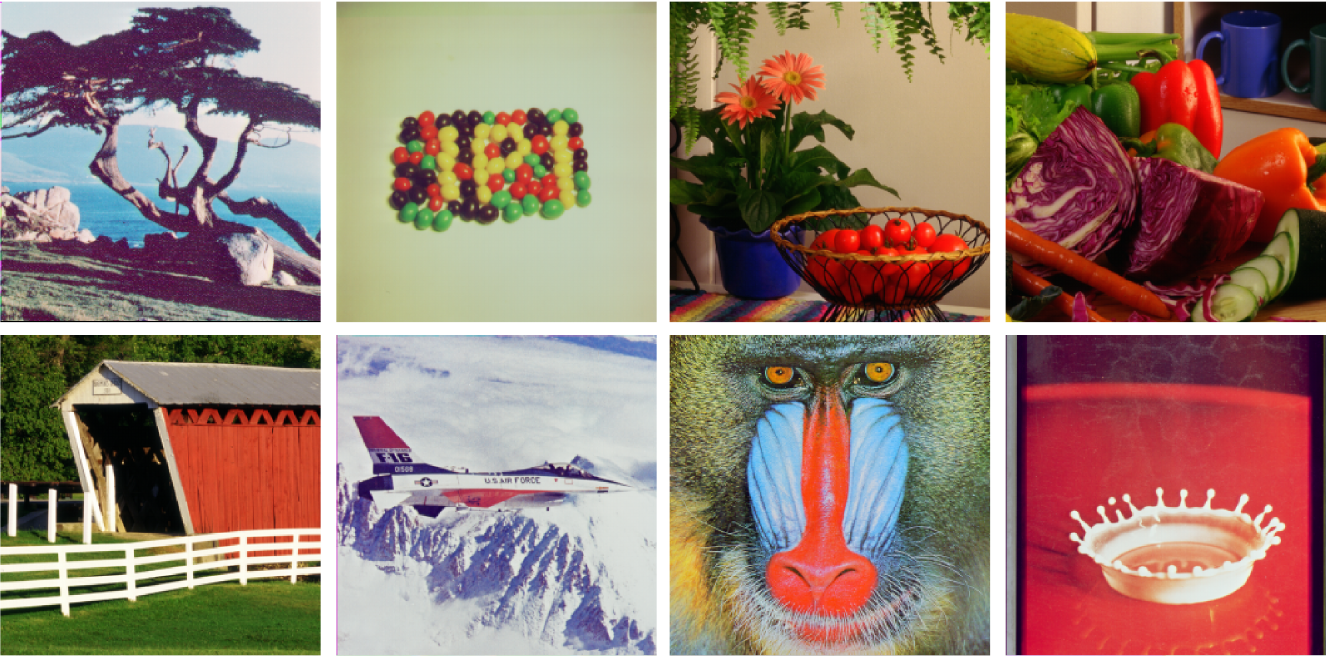}
	\caption{The $8$ color images (from left to right, and from top to bottom: $Tree, Beans, Flower, Vegetable, House, Airplane, Barbara, Splash$. They are all size $256\times 256\times 3$).}
	\label{i1}
\end{figure}

Eight benchmark color images as shown in Figure \ref{i1}, were  selected from SIPI Image Database\footnote{http://sipi.usc.edu/database/database.php}  and McMaster Dataset to demonstrate the effectiveness of the method. In order to fully demonstrate this effectiveness, 50 color images were also randomly selected from Berkeley Segmentation Dataset (BSD)\footnote{Available: https://www2.eecs.berkeley.edu/Research/Projects/CS/vision/bsds/} as further test samples. 
All the experiments were implemented in MATLAB R2019a, on a PC with a 3.00GHz CPU and 8GB RAM.

\subsubsection{Evaluation index setting}
The peak signal to noise rate (PSNR)  and the structural similarity index (SSIM)  were utilized as the relevant indices, and the best numerical results are highlighted in \textbf{bold} font. When processing image recovery with random samples, a larger Sample Rate (SR) value means more observed pixels in a given image.

\subsection{Color Image Recovery} \label{clr}
\subsubsection{Simulations with different parameters}
As simulations with different settings of the parameters ($ \beta^1, \lambda$ truncated number $r$) offer different performance levels, a range of parameters was used to test the performance of the proposed LRQR-SR algorithm, based on  recovering random sampled images from Fig. \ref{i1}.  

\begin{figure}[htbp]
	\centering
	\includegraphics[width=80mm]{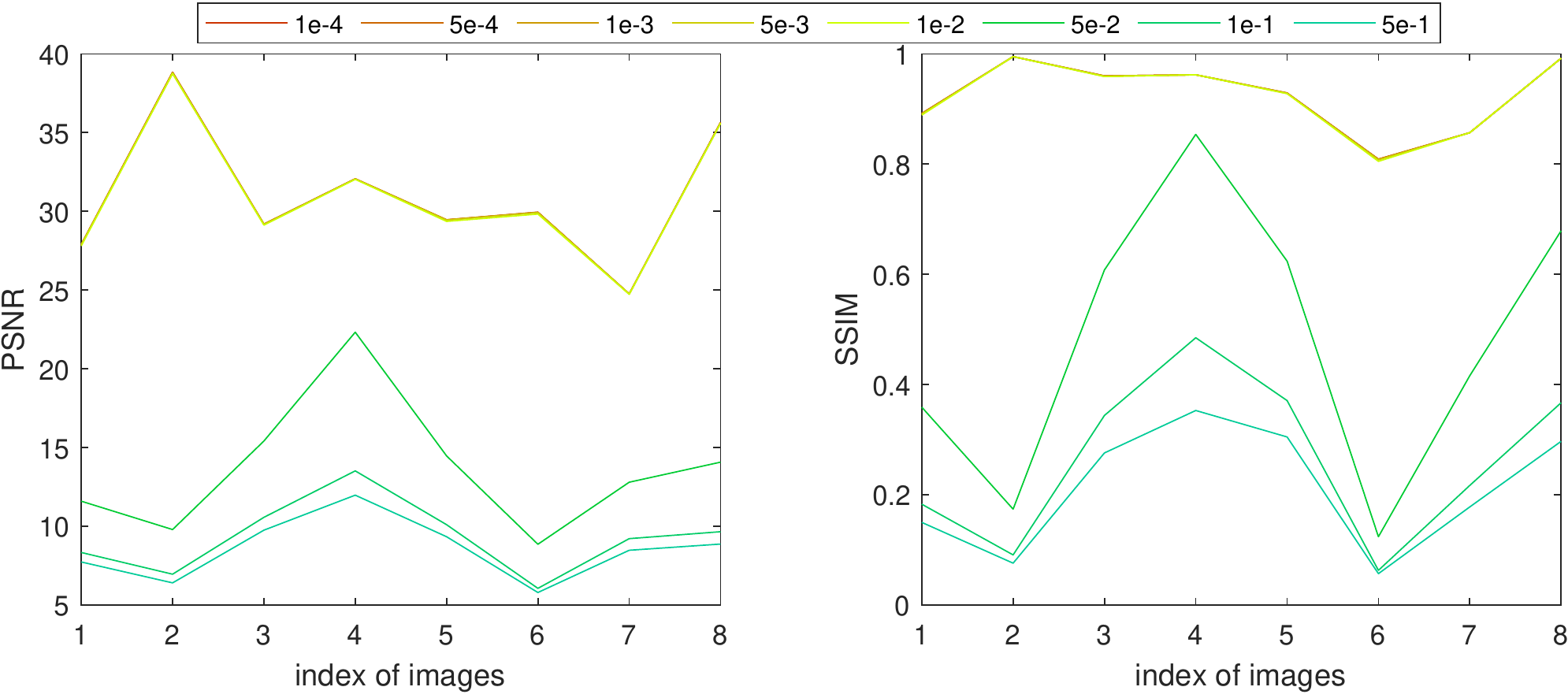}
	\caption{The PSNR and SSIM values obtained by the proposed LRQR-SR algorithm using different $\beta^1$ with other parameters fixed and $\text{SR}=0.5$. }
	\label{b0.5}
\end{figure}

\begin{figure}[htbp]
	\centering
	\includegraphics[width=80mm]{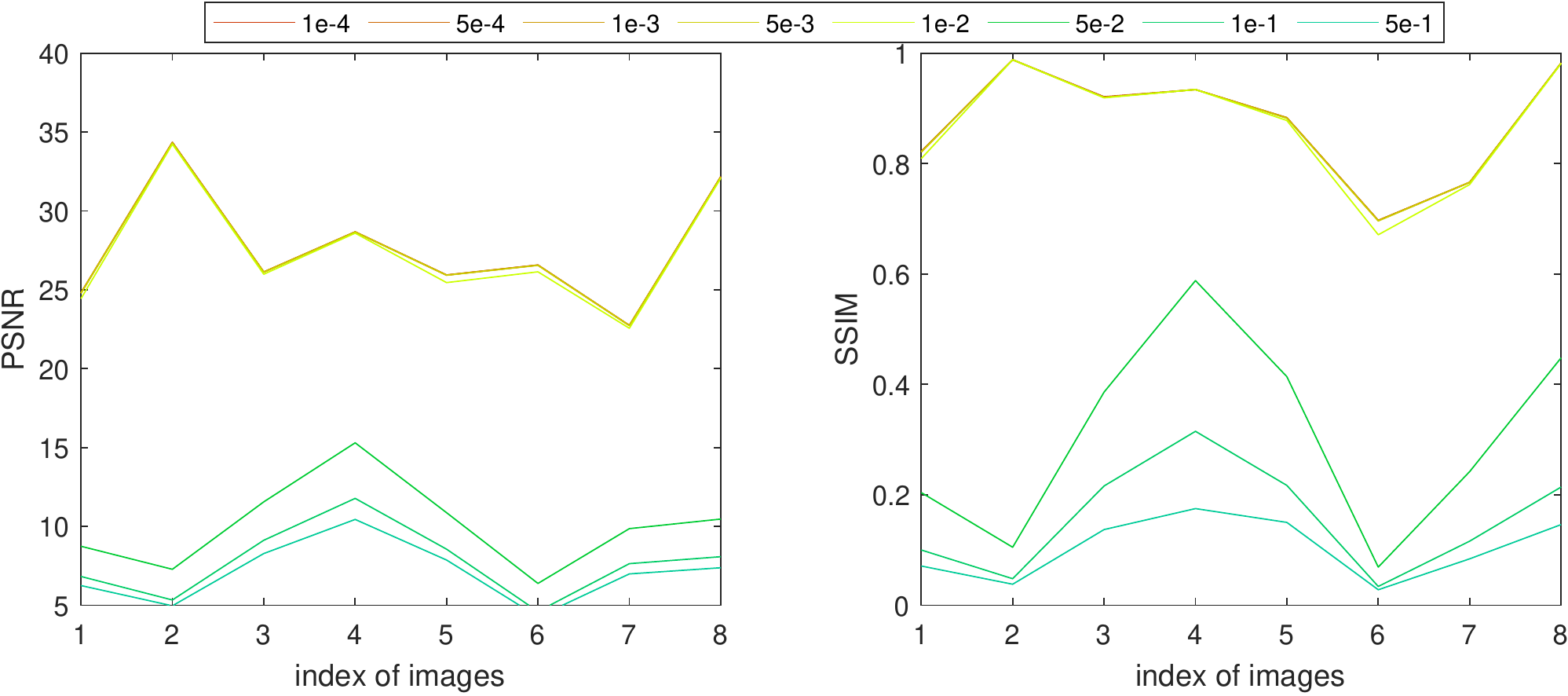}
	\caption{The PSNR and SSIM values obtained by the proposed LRQR-SR algorithm using different $\beta^1$ with other parameters fixed and $\text{SR}=0.3$.}
	\label{b0.7}
\end{figure}

\begin{figure}[htbp]
	\centering
	\includegraphics[width=80mm]{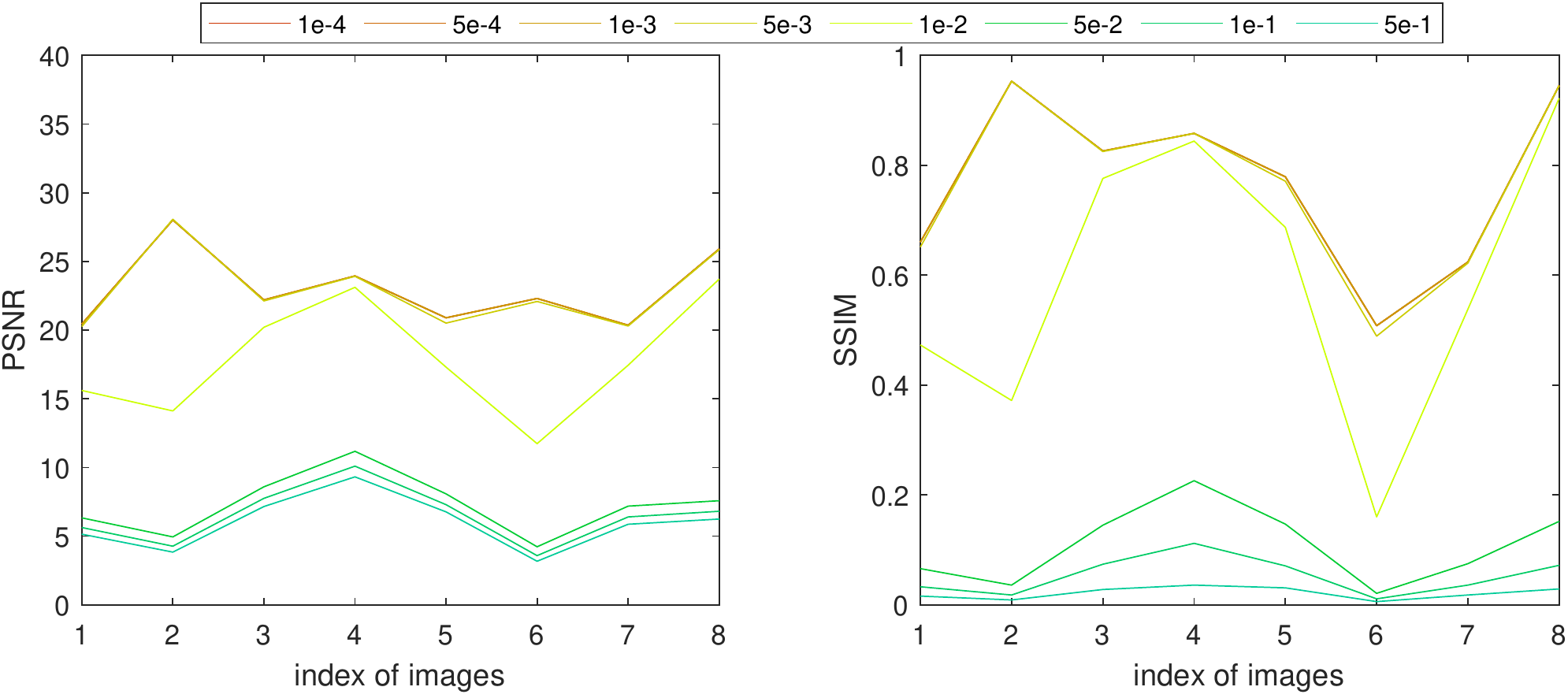}
	\caption{The PSNR and SSIM values obtained by the proposed LRQR-SR algorithm using different $\beta^1$ with other parameters fixed and $\text{SR}=0.1$.}
	\label{b0.9}
\end{figure}

The influence of different parameter values ($\beta^1=\{1e-4, 5e-4, 1e-3, 5e-3, 1e-2, 5e-2, 1e-1, 5e-1\}$) on the experimental results was first tested with the other parameters fixed ($\lambda=0.1, r=30, \rho=1.01$) and $\text{SR}=\{0.5, 0.3, 0.1\}$. The relevant PSNR and SSIM results are plotted in Fig. \ref{b0.5}-\ref{b0.9}, showing that when $\beta^1\geq1e-2$ the recovery effect is minimal. The best recovery results are instead obtained with different degrees of sampling when $\beta^1 =1e-4$.

\begin{figure}[htbp]
	\centering
	\includegraphics[width=80mm]{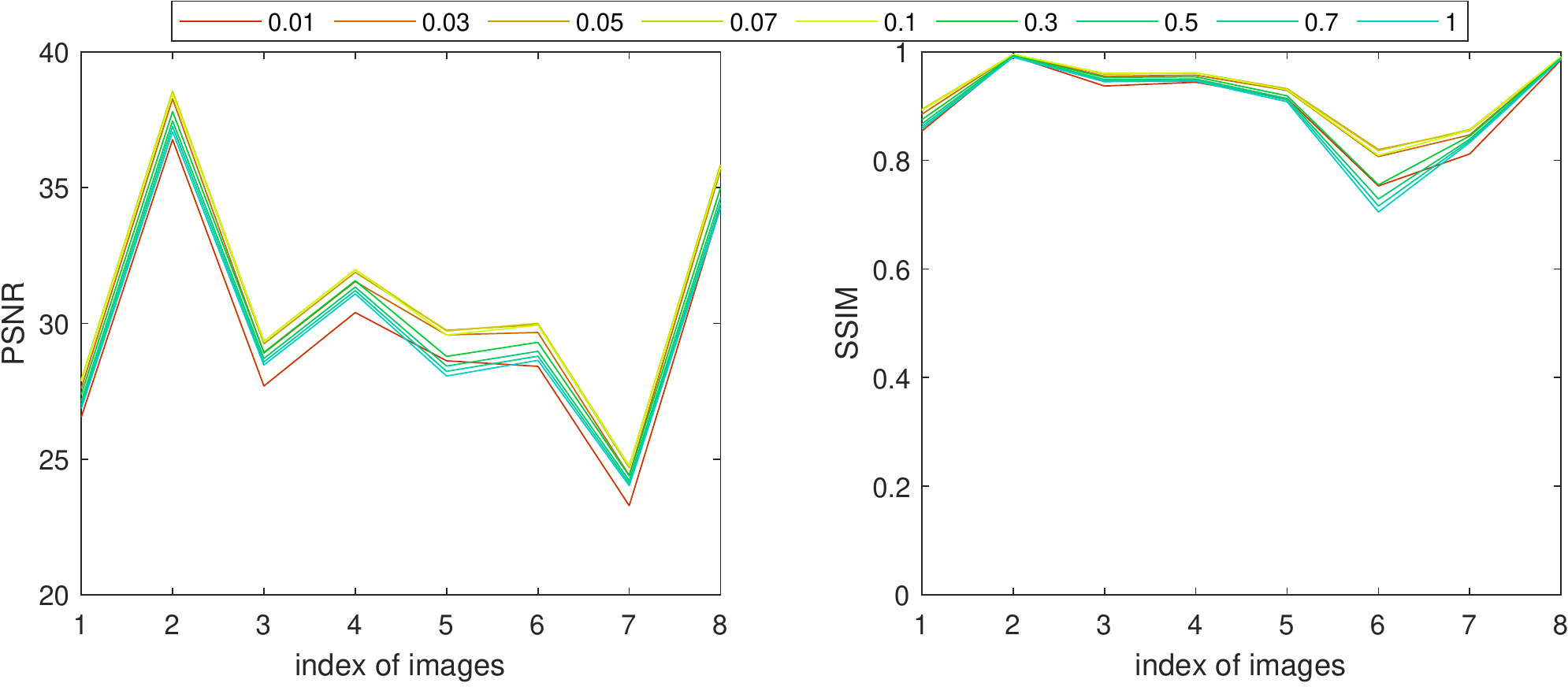}
	\caption{The PSNR and SSIM values obtained by the proposed LRQR-SR algorithm using different $\lambda$ with other parameters fixed and $\text{SR}=0.5$.}
	\label{l0.5}
\end{figure}

\begin{figure}[htbp]
	\centering
	\includegraphics[width=80mm]{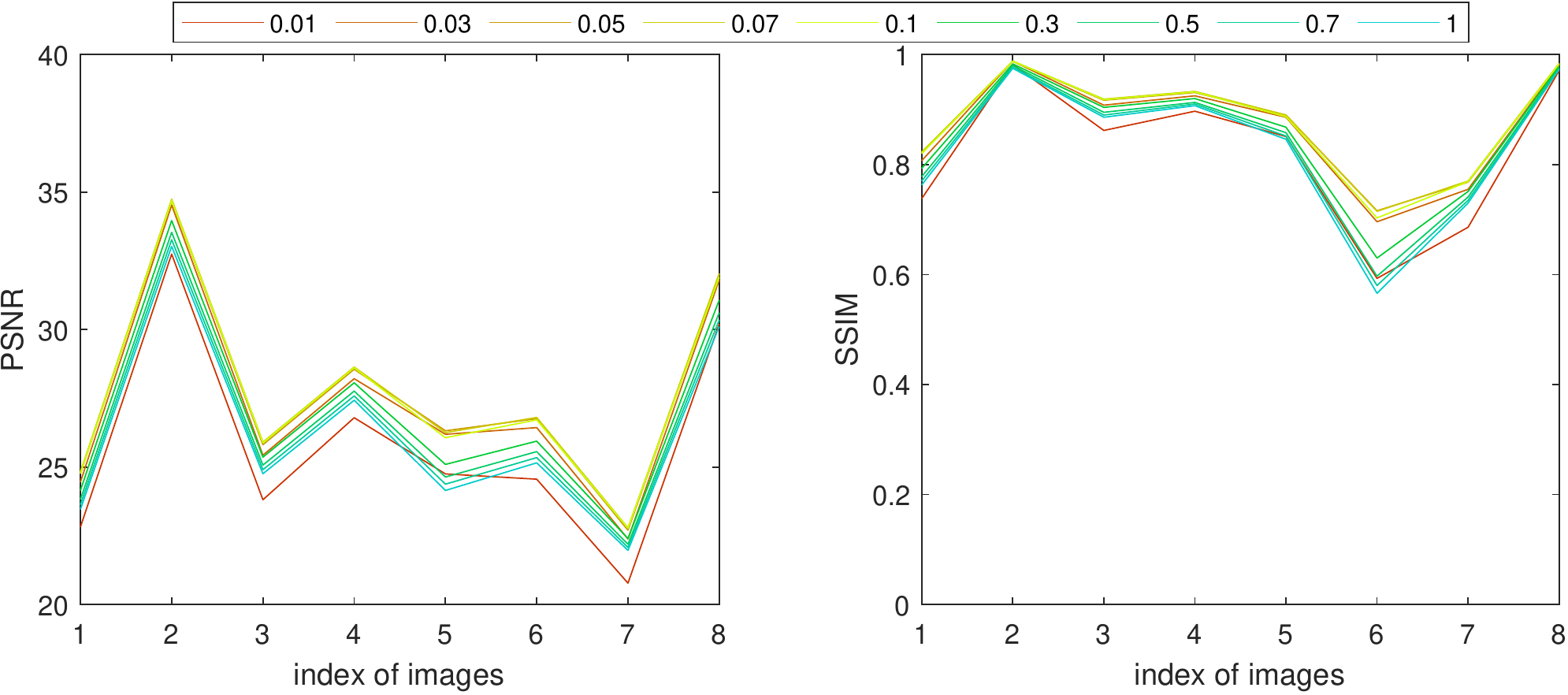}
	\caption{The PSNR and SSIM values obtained by the proposed LRQR-SR algorithm using different $\lambda$ with other parameters fixed and $\text{SR}=0.3$.}
	\label{l0.7}
\end{figure}

\begin{figure}[htbp]
	\centering
	\includegraphics[width=80mm]{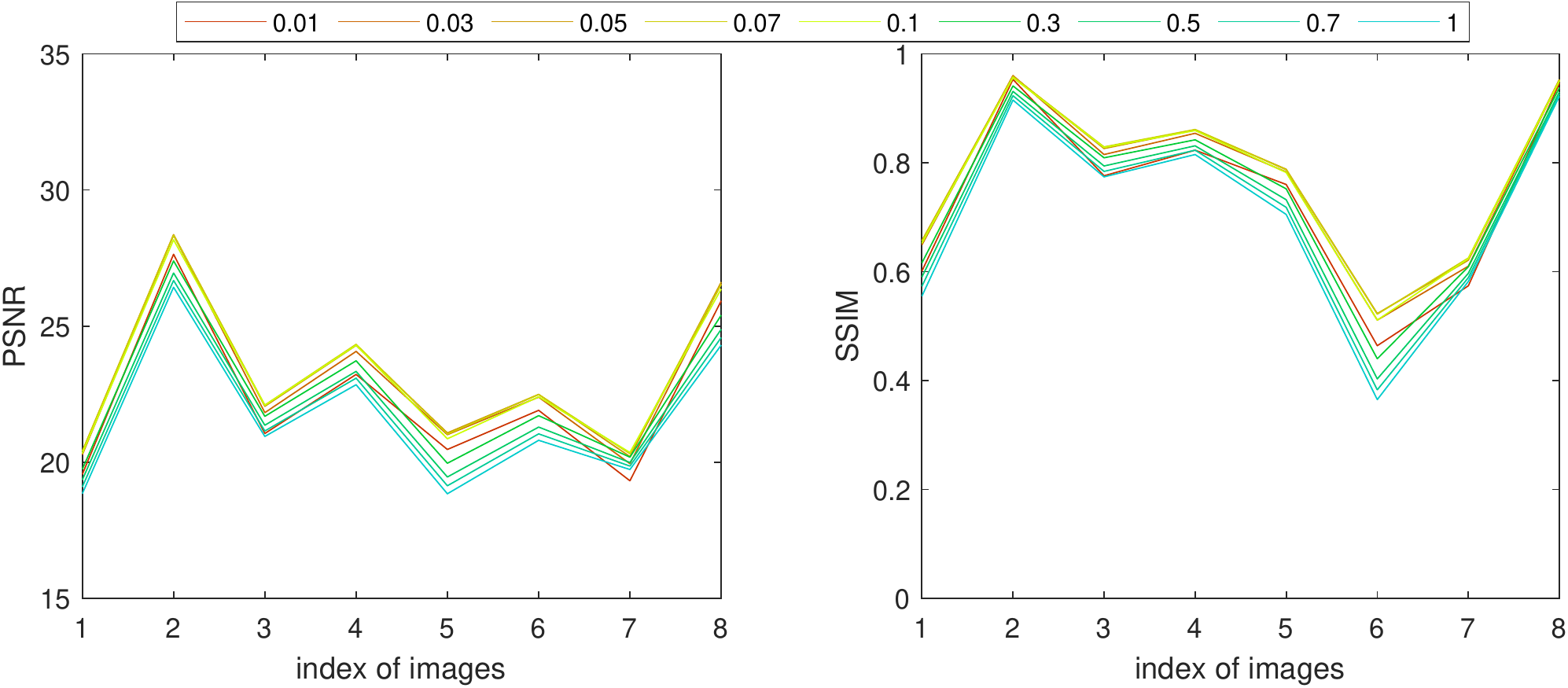}
	\caption{The PSNR and SSIM values obtained by the proposed LRQR-SR algorithm using different $\lambda$ with other parameters fixed and $\text{SR}=0.1$.}
	\label{l0.9}
\end{figure}

The effect of different parameter values ($\lambda=\{0.01, 0.03, 0.05,0.07, 0.1, 0.3, 0.5, 0.7, 1\}$) on the recovery results was then tested with other parameters fixed ($\beta^1=1e-4, r=30, \rho=1.01$) and $\text{SR}=\{0.5, , 0.3, 0.1\}$. In Fig. \ref{l0.5}-\ref{l0.9}, the relevant  PSNR and SSIM values are given, showing that if the value of $\lambda$ is too large or too small, a better recovery results cannot be obtained. The best recovery results are obtained with different SRs when $\lambda=0.07$.

\begin{figure}[htbp]
	\centering
	\includegraphics[width=80mm]{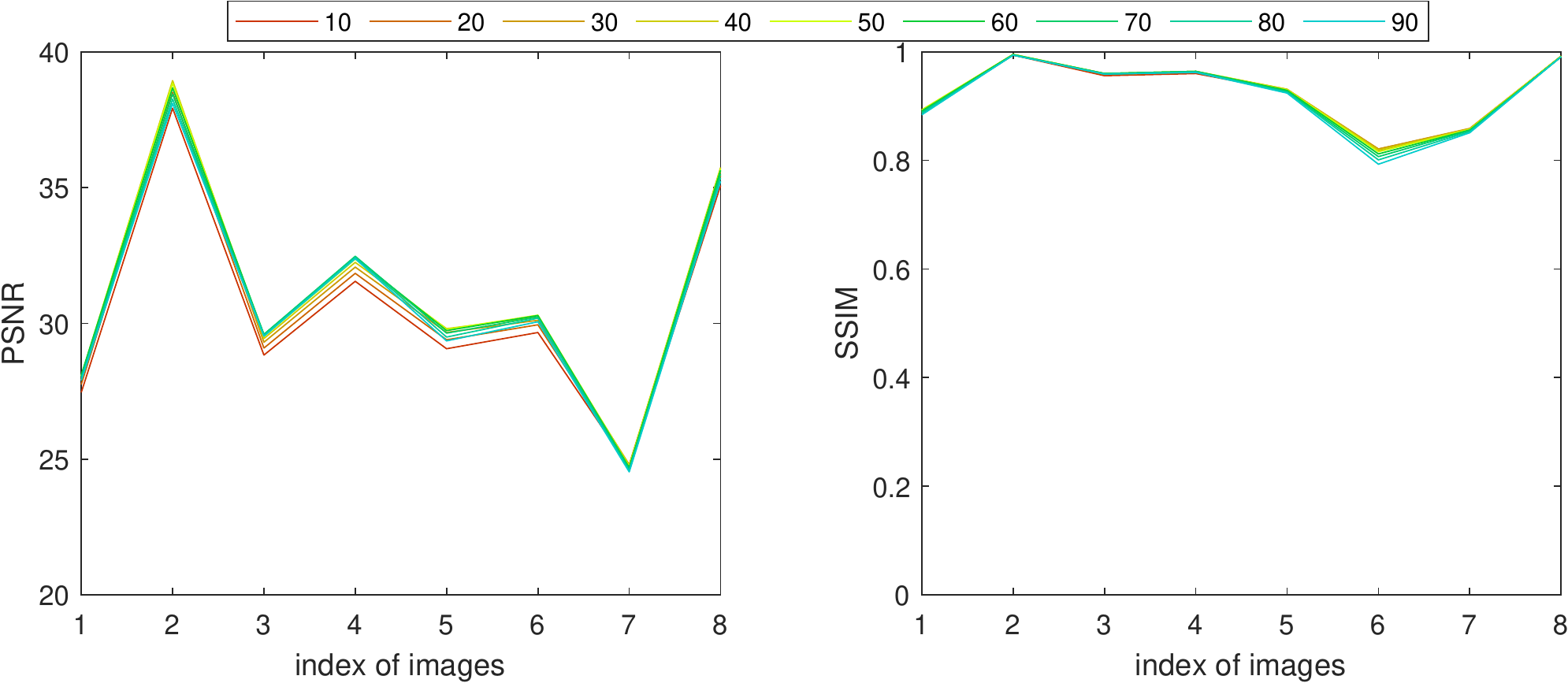}
	\caption{The PSNR and SSIM values obtained by the proposed LRQR-SR algorithm using different rank with other parameters fixed and $\text{SR}=0.5$.}
	\label{r0.5}
\end{figure}

\begin{figure}[htbp]
	\centering
	\includegraphics[width=80mm]{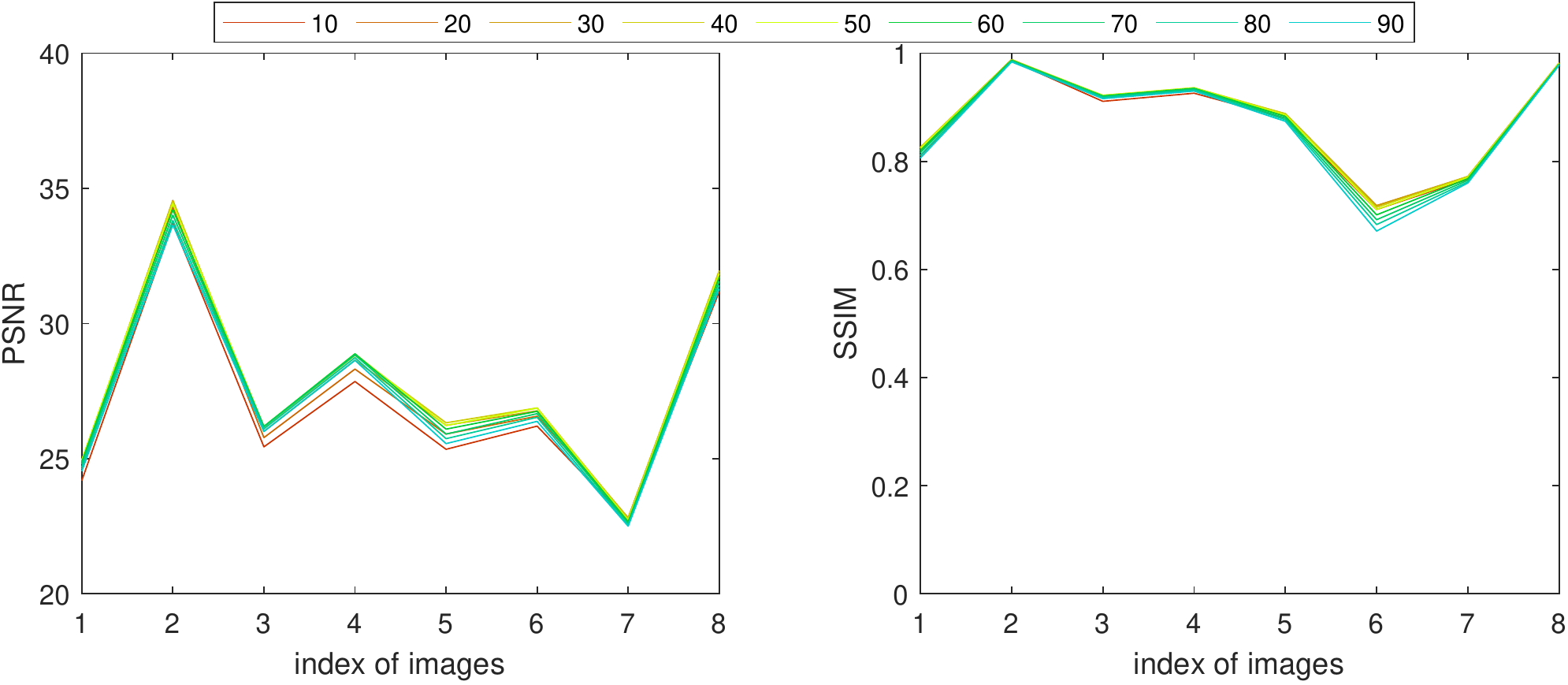}
	\caption{The PSNR and SSIM values obtained by the proposed LRQR-SR algorithm using different rank with other parameters fixed and $\text{SR}=0.3$.}
	\label{r0.7}
\end{figure}

\begin{figure}[htbp]
	\centering
	\includegraphics[width=80mm]{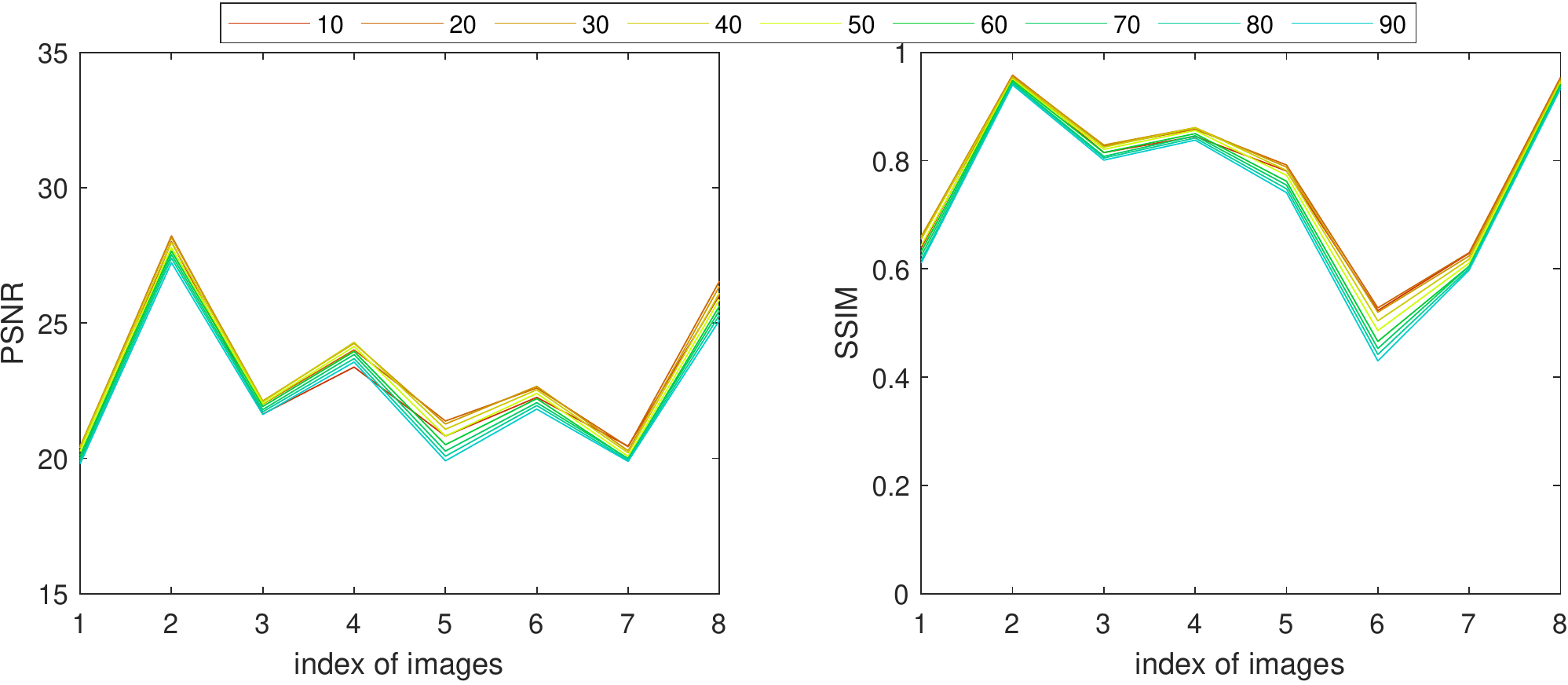}
	\caption{The PSNR and SSIM values obtained by the proposed LRQR-SR algorithm using different rank with other parameters fixed and $\text{SR}=0.1$.}
	\label{r0.9}
\end{figure}

Finally, the effect of the number of truncations on the recovery effect was verified. The truncated number r was set as $r=\{10, 20, 30, 40, 50, 60, 70, 80, 90\}$ with other parameters fixed ($\beta^1=1e-4, \lambda=0.07, \rho=1.01$) and $\text{SR}=\{0.5, 0.3, 0.1\}$. In Fig. \ref{r0.5}-\ref{r0.9} show the relevant PSNR and SSIM values are given, indicating that the best recovery results were obtained when the degree of sampling was high ($\text{SR}=0.5$) and the truncated number $r=\{40, 50\}$. However, when the degree of sampling is lower ($\text{SR}=\{0.3, 0.1\}$) and $r=\{30, 40\}$, good recovery results are also obtained. This is also consistent with the fact that the more missing pixels in the observed image, the more low-rank constraints are required to improve the recovery effect. Intuitively, when the observed image is missing a lot of pixels, the truncation would contains less useful information.
\subsubsection{Images recovery with random sample}
The LRQR-SR algorithm was compared with several other methods mentioned previously by setting $\text{SR}=\{0.3, 0.2, 0.1\}$. The parameters of  LRQR-SR were set as $\beta^1=1e-4, \lambda=0.07, \rho=1.01$, while the truncation number $r=\{40, 30, 20\}$ was decided by the SR: the lower the SR, the less truncation is required. 

\begin{figure*} [htbp]
	\centering
	\includegraphics[width=180mm]{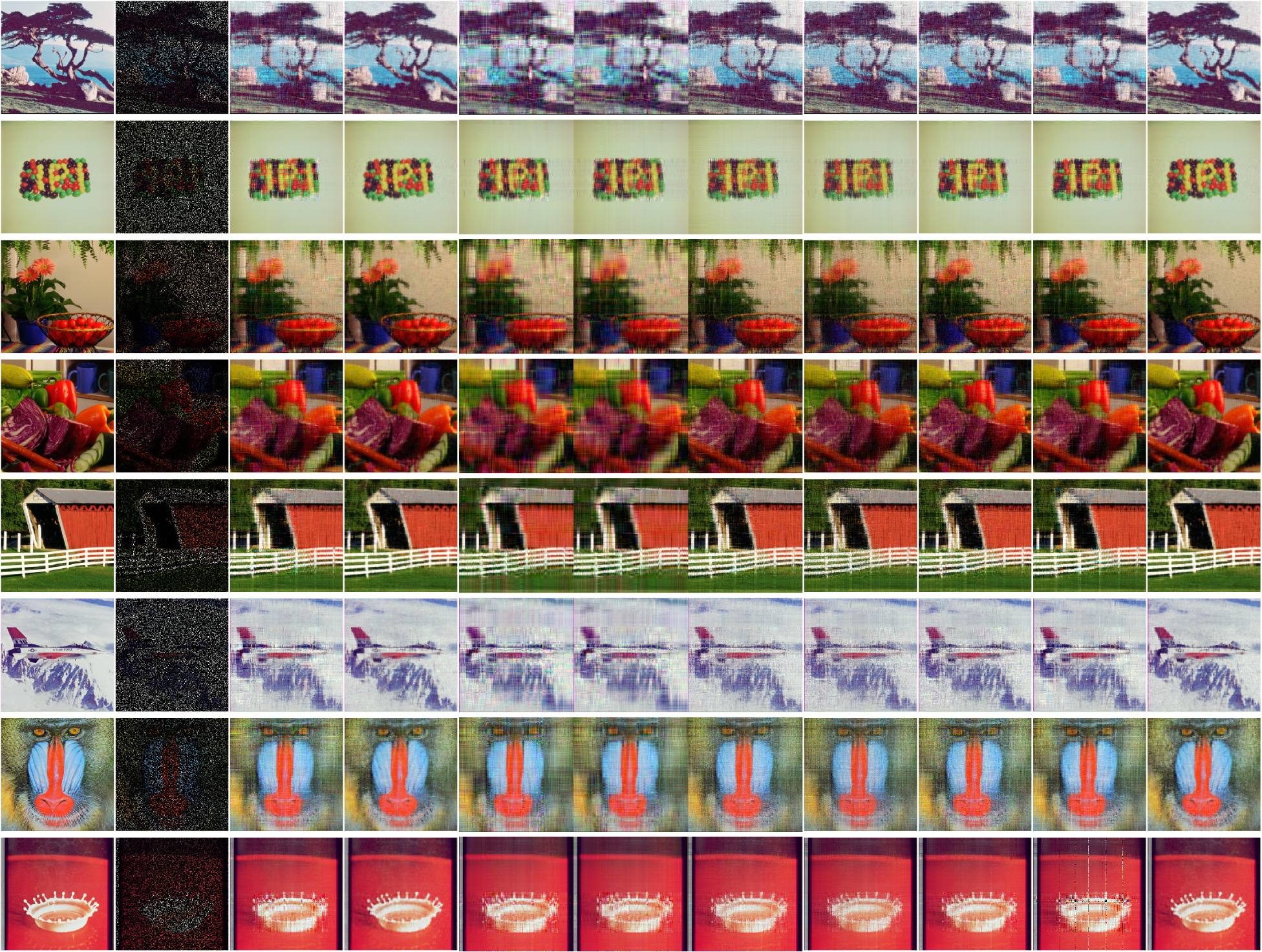}
	\caption{The first column is the original image and the second column is the observed image (SR=0.2). The 3-\textit{th} column to 11-\textit{th} column are the completion results of TNNR, TNN-SR, D-N, F-N,LRQA, Q-FFN, Q-DNN, QTNN, and LRQR-SR, respectively. The corresponding parameters are summarized in Table \ref{t1}.}
	\label{0.8}
\end{figure*}

Fig. \ref{0.8} displays the visual comparisons between the designed novel LRQR-SR method and the other methods of comparison for the eight tested color images when $\text{SR}=0.2$. The PSNR and SSIM results of for recovery as seen in  Fig. \ref{i1} with $\text{SR}=\{0.3, 0.2, 0.1\}$ are presented in Table \ref{t1}. As shown, as compared with other methods, across all SR values, the results obtained by D-N and F-N do not show particularly clear recovery. However, such factorization skills are more effective when operated in the quaternion domain, as in Q-FFN and Q-DNN. The validity of quaternion-based methods is thus illustrated by these results. The results for TNNR are also inferior to those of TNNR-SR, highlighting that only utilizing low-rankness as prior is insufficient to recover an image more accurately. This supports the reasoning behind introducing sparsity to the LRQR-SR method. 

In comparison with the other options, the developed  LRQR-SR method also provides the most visually optimal results, with crisp details. As shown in the data presented in TABLE \ref{t1}, when the value of SR is very low, utilizing only the truncated skill cannot recover the potential images accurately, while  LRQR-SR uniformly outperforms its comparators in terms of both PSNR and SSIM values. 

\begin{figure} [htbp]
	\centering
	\includegraphics[width=90mm]{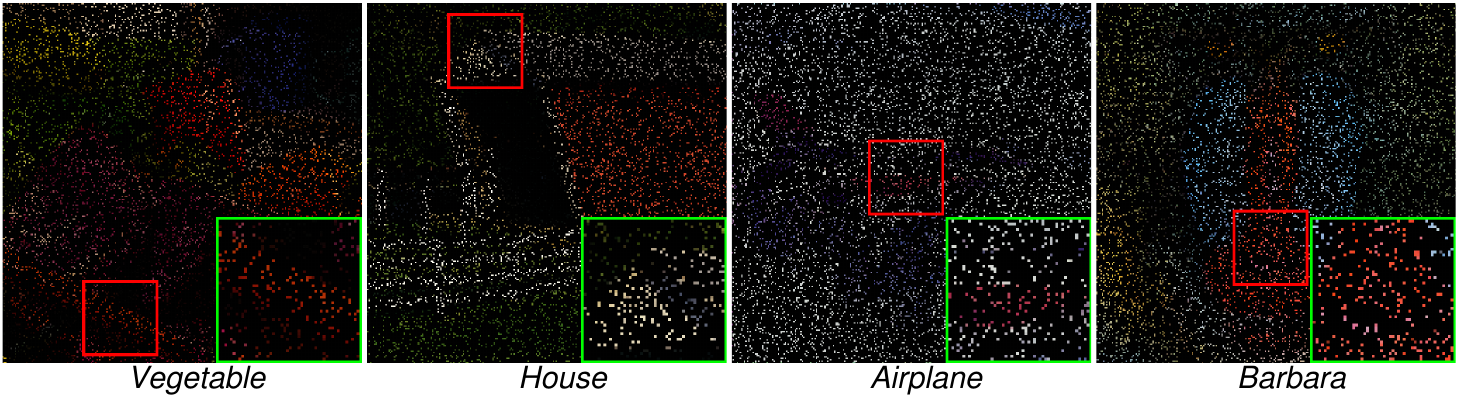}
	\caption{The enlargement of observed images under SR = 0.1, corresponding to Fig. \ref{z1}-\ref{z4}.}
	\label{z0}
\end{figure}

\begin{figure} [htbp]
	\centering
	\includegraphics[width=90mm]{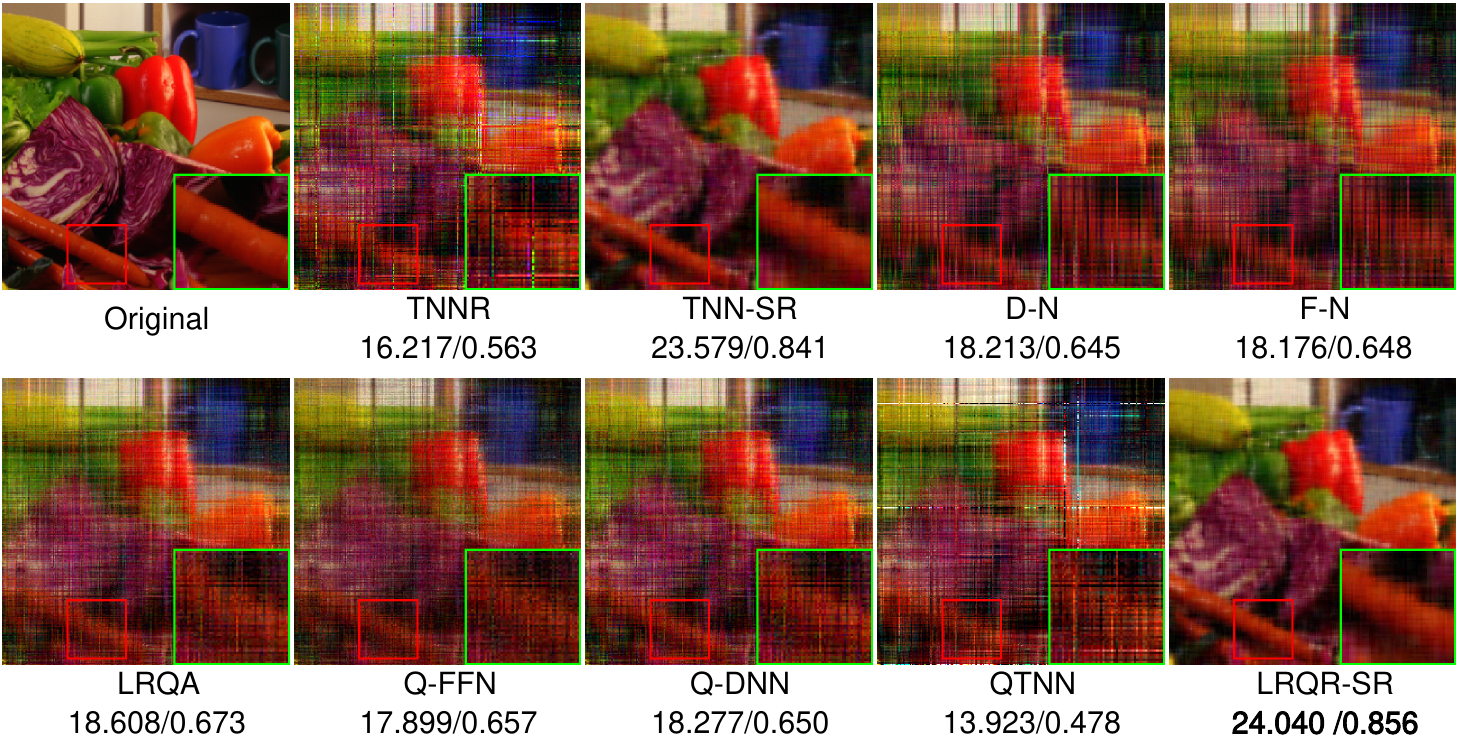}
	\caption{The recovered results of image $Vegetable$ by different methods under SR = 0.1.}
	\label{z1}
\end{figure}

\begin{figure} [htbp]
	\centering
	\includegraphics[width=90mm]{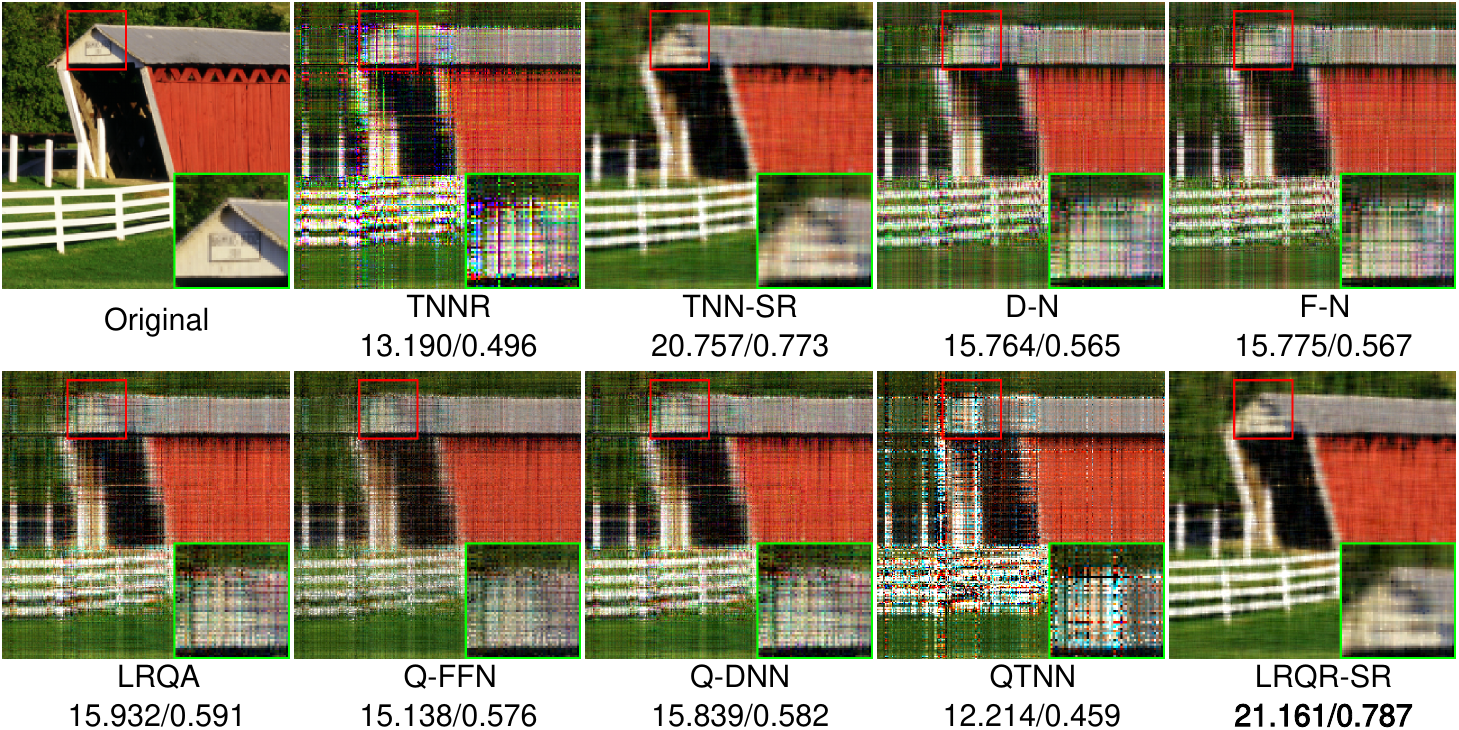}
	\caption{The recovered results of image $House$ by different methods under SR = 0.1.}
	\label{z2}
\end{figure}

\begin{figure} [htbp]
	\centering
	\includegraphics[width=90mm]{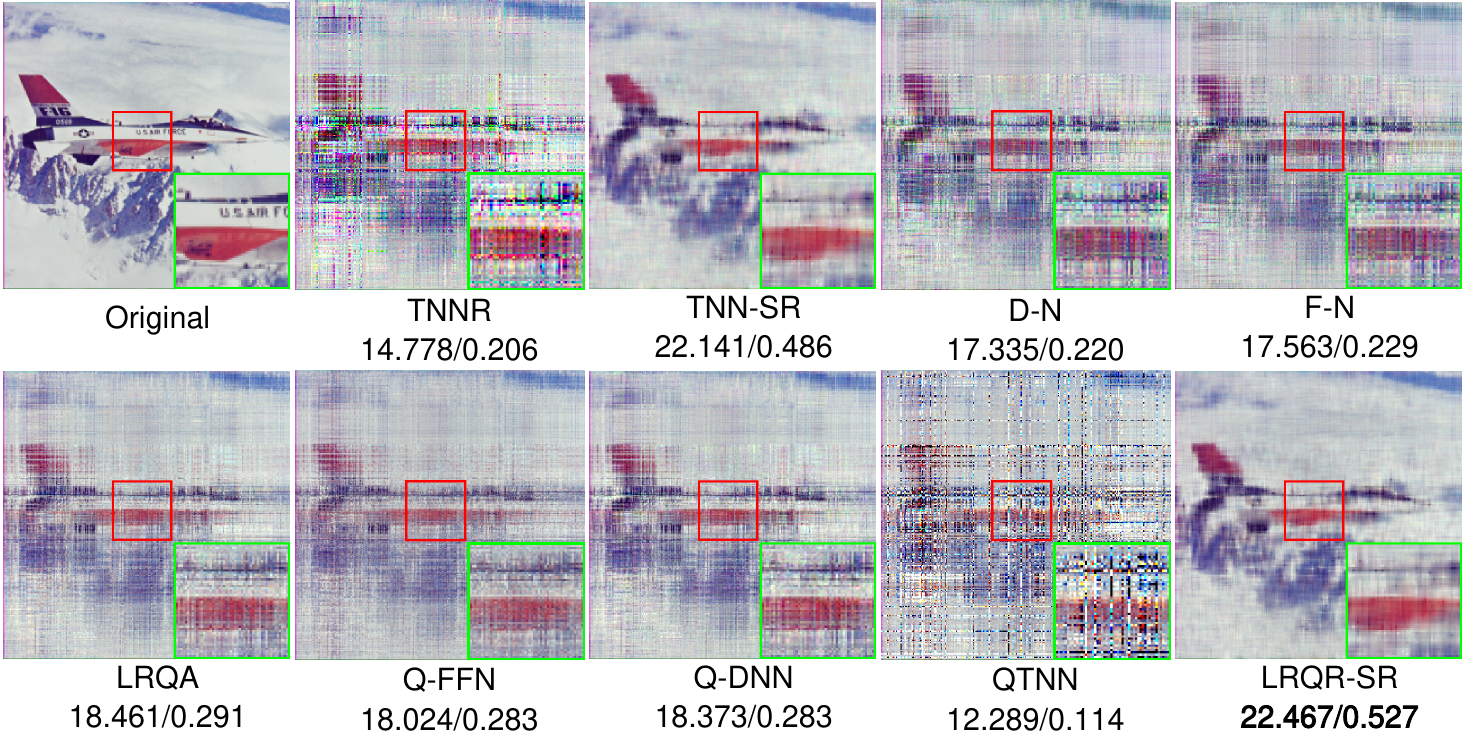}
	\caption{The recovered results of image $Airplane$ by different methods under SR = 0.1.}
	\label{z3}
\end{figure}

\begin{figure} [htbp]
	\centering
	\includegraphics[width=90mm]{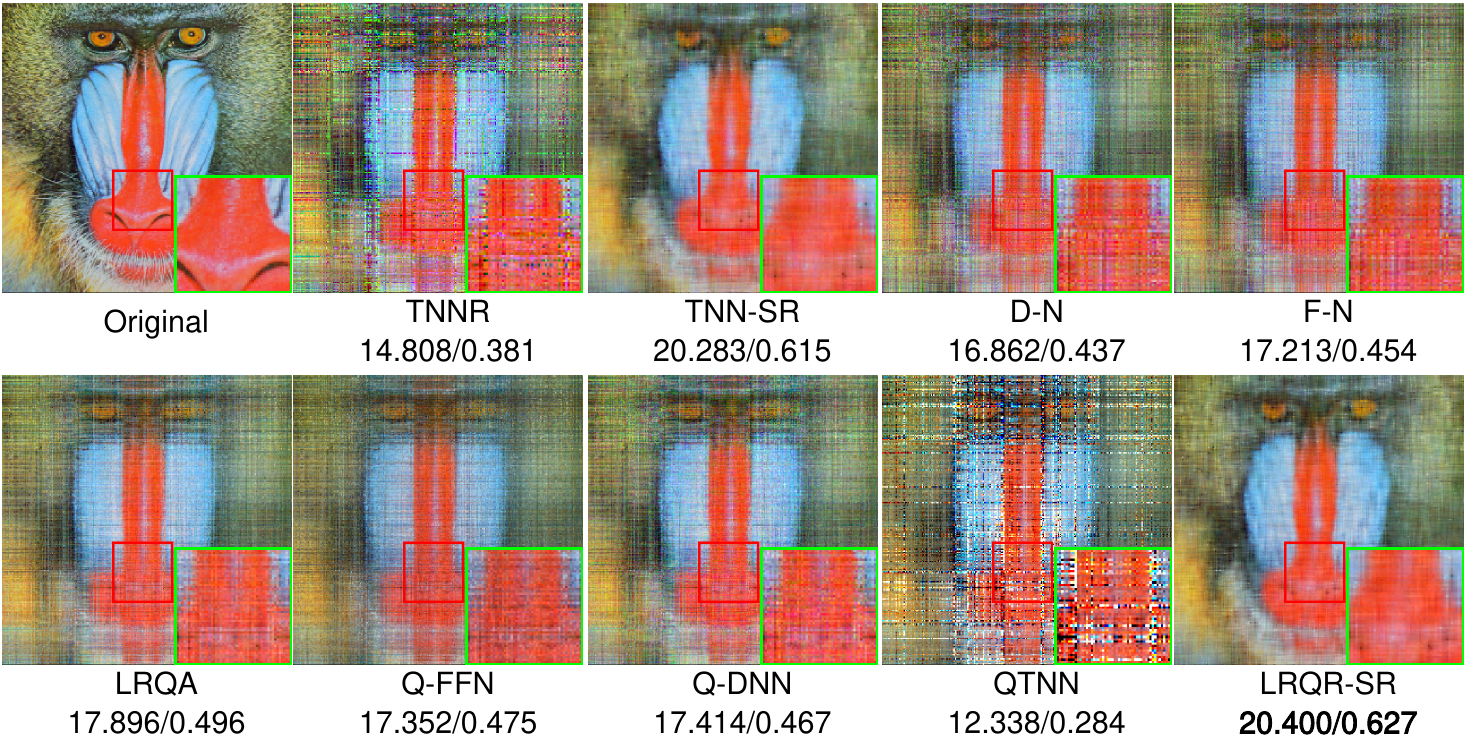}
	\caption{The recovered results of image $Barbara$ by different methods under SR = 0.1.}
	\label{z4}
\end{figure}

Fig. \ref{z1}-\ref{z4} compares the visual results for all the competing completion approaches related to recovering $Vegetable, House, Airplane, Barbara$ under SR = 0.1. The corresponding observed images are shown in Fig. \ref{z0}. The image in the green box is the zoomed-in image of that in the red box. These images show that restricting recovery only to low rankness may lose some local details. Moreover, the TNNR and QTNN approaches restore the image only roughly when the sample rate is low, with other methods  suffering from similar problems. Comparing just the two optimal algorithms (TNN-SR and LRQR-SR), although the effect gap is not visually apparent across the restored images, the corresponding PSNR and SSIM prove the superiority of the proposed method.
  \renewcommand{\arraystretch}{1.2} 
\begin{table*}[htbp] 
	\centering  
	\caption{The PSNR/SSIM values obtained by different recovery algorithms for 8 color images}  
	\label{t1}  
	\scalebox{0.85}{
	\begin{tabular}{|c|c|c|c|c|c|c|c|c|c|}  
		\hline  
		\multirow{2}{*}{Image}&  
		\multicolumn{9}{c|}{$\text{SR}=0.3$}\cr\cline{2-10}  
		&TNNR\cite{DBLP:journals/pami/HuZYLH13}&TNN-SR\cite{DBLP:journals/spic/DongXGHW18}&D-N\cite{DBLP:journals/pami/ShangCLLL18}& F-N\cite{DBLP:journals/pami/ShangCLLL18}&LRQA \cite{DBLP:journals/tip/ChenXZ20}& Q-FFN\cite{DBLP:journals/tsp/MiaoK20}& Q-DNN\cite{DBLP:journals/tsp/MiaoK20}& QTNN \cite{DBLP:journals/jvcir/YangKM21}& LRQR-SR\cr  
		\hline  
		 $Tree$&20.936/0.670  &23.275/0.793  &18.356/0.490 &18.309/0.485  &21.148/0.668 &20.600/0.650 &21.070/0.657 &21.028/0.676  &\pmb{24.915/0.827}  \cr\hline 
		 $Beans$&27.699/0.954 &33.799/0.984 & 26.509/0.938 &26.559/0.938 &27.950/0.950 &28.100/0.957 &28.133/0.955 & 28.324/0.960&\pmb{34.583/0.988} \cr\hline   
		$Flower$&22.275/0.818&25.343/0.903&20.061/0.735&20.093/0.738  &22.353/0.816 &22.053/0.796 &22.260/0.807&22.588/0.830&\pmb{26.128/0.920}\cr\hline    
		$Vegetable$&24.053/0.848&27.861/0.919&20.871/0.757&20.861/0.758 &24.294/0.854&24.002/0.852&24.311/0.848&24.359/0.859&\pmb{28.865/0.936}\cr\hline  
		$House$&21.838/0.802&25.357/0.873&19.299/0.711&19.288/0.712  &22.162/0.804&21.786/0.786&22.281/0.799&21.866/0.806&\pmb{26.272/0.887}\cr\hline  
		$Airplane$&23.182/0.544&26.237/0.676&21.028/0.400&21.078/0.399&	23.101/0.527&22.777/0.526&23.059/0.518 &23.214/0.548&\pmb{26.851/0.712}\cr\hline
		 $Barbara$&20.853/0.690&22.525/0.758&19.992/0.590&20.038/0.591  &20.688/0.679&19.433/0.603&20.065/0.643&21.046/0.697&\pmb{22.734/0.769}\cr\hline
		$Splash$&26.761/0.952 &31.508/0.979 &24.838/0.933 &24.938/0.934 &26.926/0.951&26.699/0.950&27.124/0.951&26.860/0.952&\pmb{32.126/ 0.982}\cr  
		\hline  
		\hline  
		\multirow{2}{*}{Image}&  
		\multicolumn{9}{c|}{$\text{SR}=0.2$}\cr\cline{2-10}  
		&TNNR&TNN-SR&D-N& F-N&LRQA & Q-FFN& Q-DNN& QTNN& LRQR-SR\cr \hline  
		$Tree$&18.517/0.545&21.921/0.730&17.465/0.433&17.551/0.447  &18.910/0.551&18.354/0.515&18.554/0.523&18.583/0.548
		&\pmb{23.093/0.769}\cr\hline 
		$Beans$&24.163/0.918&31.165/0.974&24.846/0.918&24.990/0.918
		&25.103/0.924&25.018/0.926&25.197/0.927&25.033/0.930
		&\pmb{31.804/0.979}\cr\hline   
		$Flower$&20.112/0.736&23.599/0.866&19.214/0.692&19.310/0.698   &20.442/0.745&20.068/0.725&20.062/0.719&20.465/0.753
		&\pmb{24.138/0.884}\cr\hline    
		$Vegetable$&21.674/0.781&25.978/0.891&20.299/0.732&20.299/0.735  &21.947/0.791&21.902/0.790&21.737/0.777&22.028/0.796
		&\pmb{26.802/0.909}\cr\hline  
		$House$&19.232/0.723&23.615/0.839&18.423/0.677&18.397/0.677 &19.800/0.736&18.950/0.710&19.464/0.716&19.120/0.727   &\pmb{24.349/0.855}\cr\hline  	
	   $Airplane$&20.601/0.419&24.496/0.600&19.641/0.332&19.774/0.340 &20.898/0.421&20.401/0.400&20.608/0.404&20.739/0.427  &\pmb{25.002/0.636}\cr\hline	    
	 $Barbara$&19.335/0.602&21.487/0.697&18.935/0.547&19.174/0.554   &19.503/0.601&19.301/0.593&18.884/0.567&19.522/0.609
	 &\pmb{21.695/0.710}\cr\hline
	$Splash$&24.285/0.929&29.473/0.970&23.620/0.919&23.689/0.920
	&24.643/0.931&24.211/0.929&24.643/0.929&23.946/0.924  
	&\pmb{29.915/0.973}\cr  
		\hline  
		\hline  \multirow{2}{*}{Image}&  
		\multicolumn{9}{c|}{$\text{SR}=0.1$}\cr\cline{2-10}  
	     &TNNR&TNN-SR&D-N& F-N&LRQA & Q-FFN& Q-DNN& QTNN& LRQR-SR\cr \hline  
		$Tree$&13.794/0.256&19.943/0.629&14.905/0.266 &15.232/0.289
		&15.893/0.364&15.474/0.359&15.604/0.338&11.508/0.178
		&\pmb{20.435/0.657}\cr\hline 
  	  $Beans$&17.403/0.779&28.089/0.952&21.672/0.845&21.777/0.846
  	  &22.191/0.879&22.293/0.869&22.135/0.879&14.141/0.388
  	  &\pmb{28.511/0.960}\cr\hline   
	$Flower$&14.956/0.467&21.673/0.807&16.960/0.564&17.195/0.583   
	 &17.973/0.632&17.457/0.615&17.541/0.595&12.518/0.346
	 &\pmb{22.013/0.825}   \cr\hline   	 
	$Vegetable$&16.217/0.563&23.579/0.841&18.213/0.645&18.176/0.648            &18.608/0.673&17.899/0.657&18.277/0.650&13.923/0.478
	 &\pmb{24.040 /0.856}  \cr\hline  
	$House$&13.190/0.496&20.757/0.773&15.764/0.565&15.775/0.567
	&15.932/0.591&15.138/0.576&15.839/0.582&12.214/0.459
	&\pmb{21.161/0.787}\cr\hline  
	$Airplane$&14.778/0.206&22.141/0.486&17.335/0.220&17.563/0.229
	 &18.461/0.291&18.024/0.283&18.373/0.283&12.289/0.114
	 &\pmb{22.467/0.527}  \cr\hline
	$Barbara$&14.808/0.381&20.283/0.615&16.862/0.437 &17.213/0.454  &17.896/0.496&17.352/0.475&17.414/0.467&12.338/0.284
	&\pmb{20.400/0.627}   \cr\hline
	$Splash$&16.765/0.817&26.556/0.948&20.228/0.884&20.299/0.886
	&21.157/0.891&20.616/0.890&21.015/0.885&13.311/0.737
	&\pmb{26.900/0.953}\cr \hline  
	\end{tabular}}
\end{table*}  


\subsubsection{Images recovery under text mask}
The LRQR-SR algorithm was then compared with other methods under a text mask. The parameters of  LRQR-SR were set to $\beta^1=1e-4, \lambda=0.07, \rho=1.01$, and the truncated number $r=30$. 

\begin{figure} [htbp]
	\centering
	\includegraphics[width=90mm]{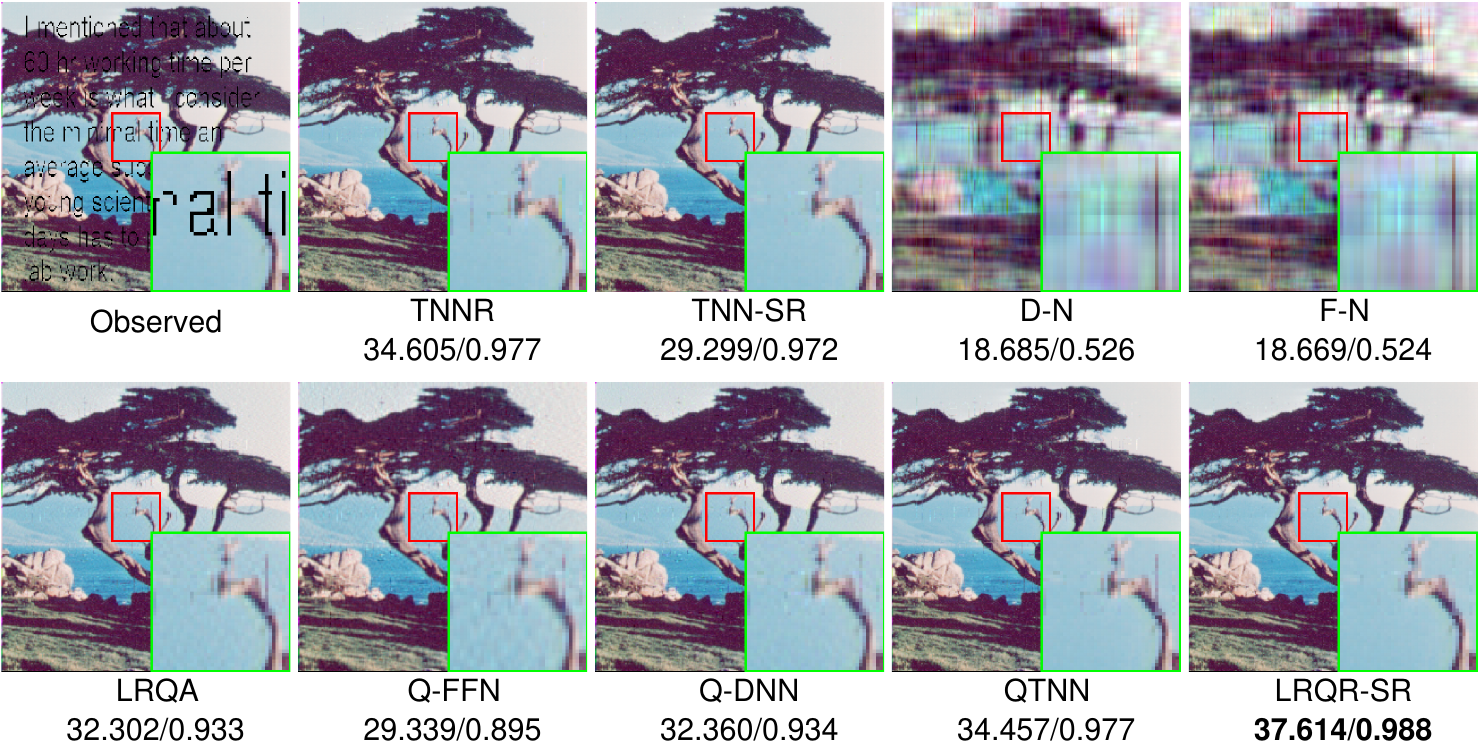}
	\caption{The recovered results of image $Tree$ by different methods under text mask}
	\label{z5}
\end{figure}

\begin{figure} [htbp]
	\centering
	\includegraphics[width=90mm]{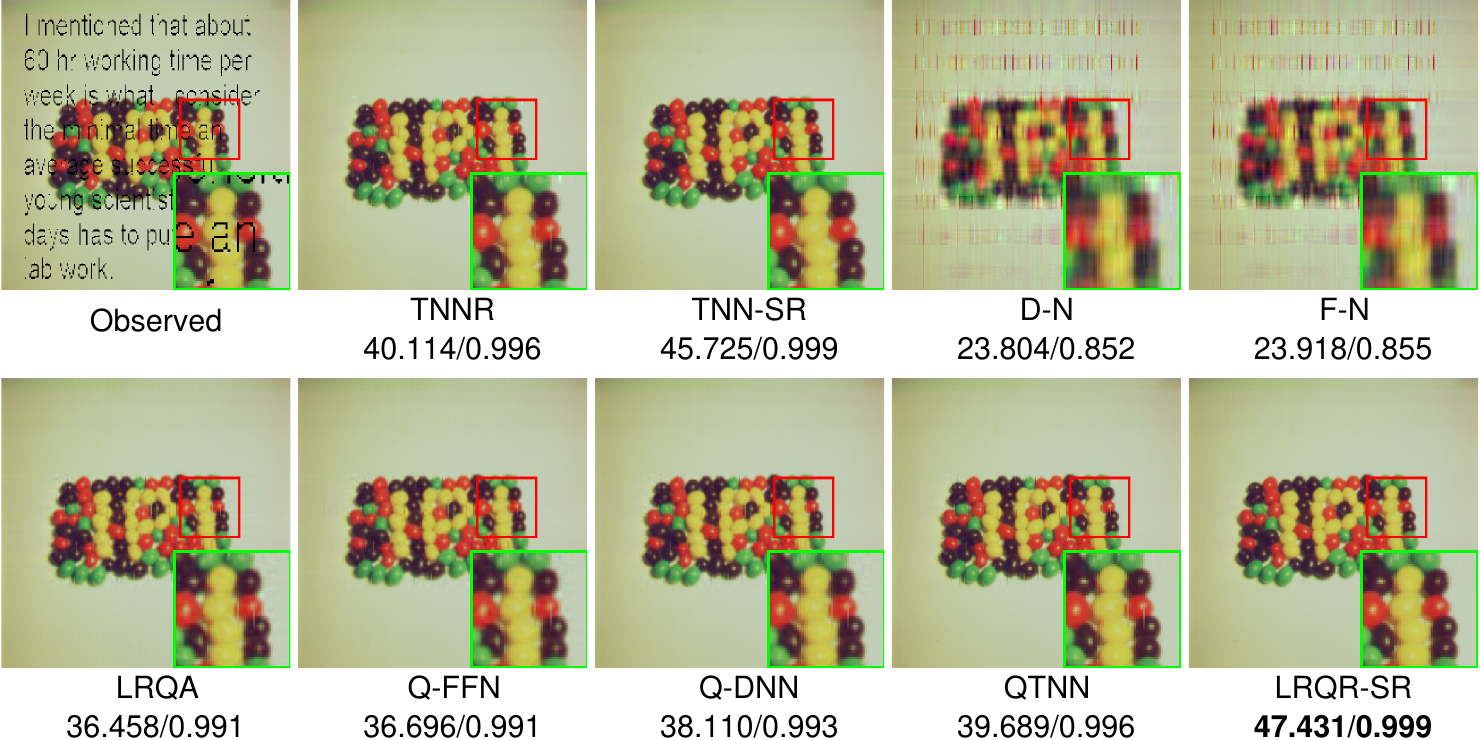}
	\caption{The recovered results of image $Beans$ by different methods under under text mask}
	\label{z6}
\end{figure}

\begin{figure} [htbp]
	\centering
	\includegraphics[width=90mm]{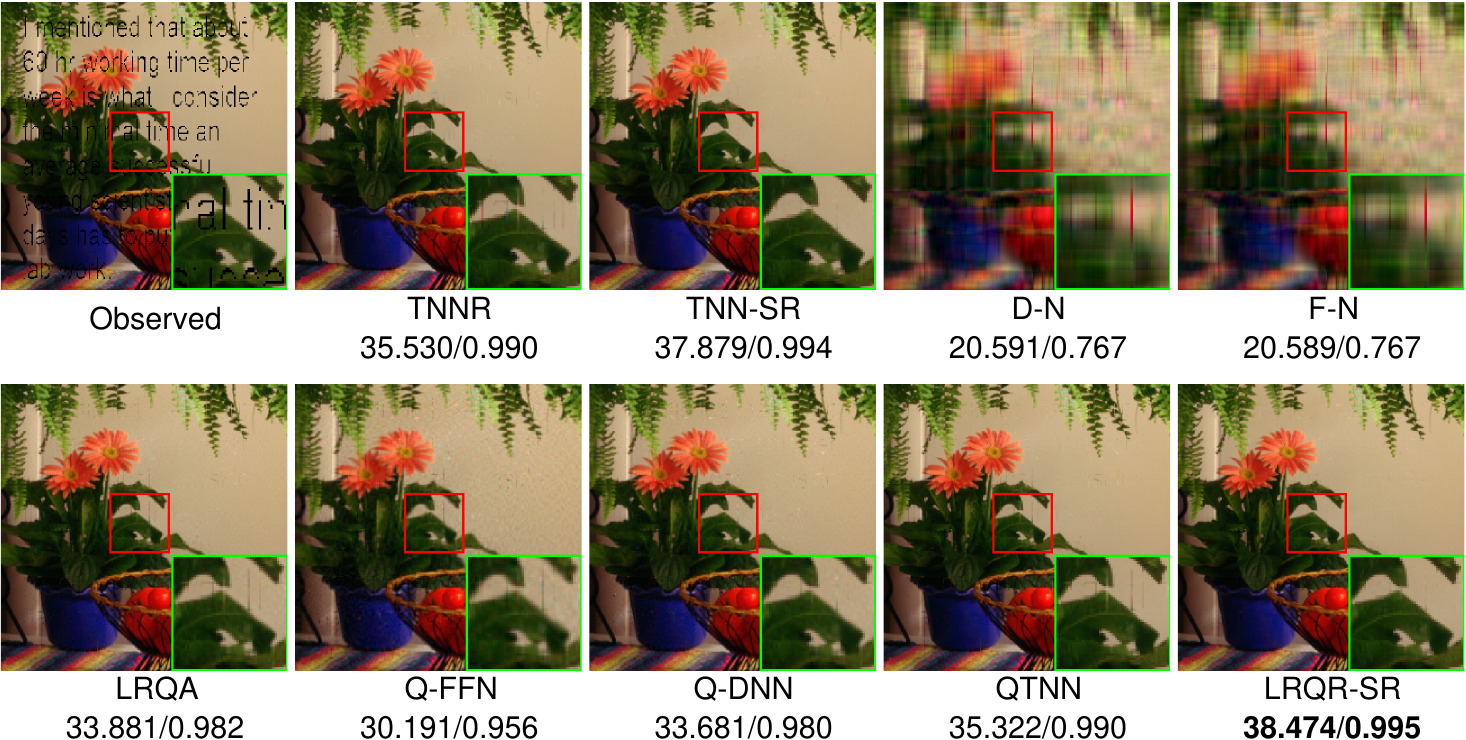}
	\caption{The recovered results of image $Vegetable$ by different methods under text mask}
	\label{z7}
\end{figure}

\begin{figure} [htbp]
	\centering
	\includegraphics[width=90mm]{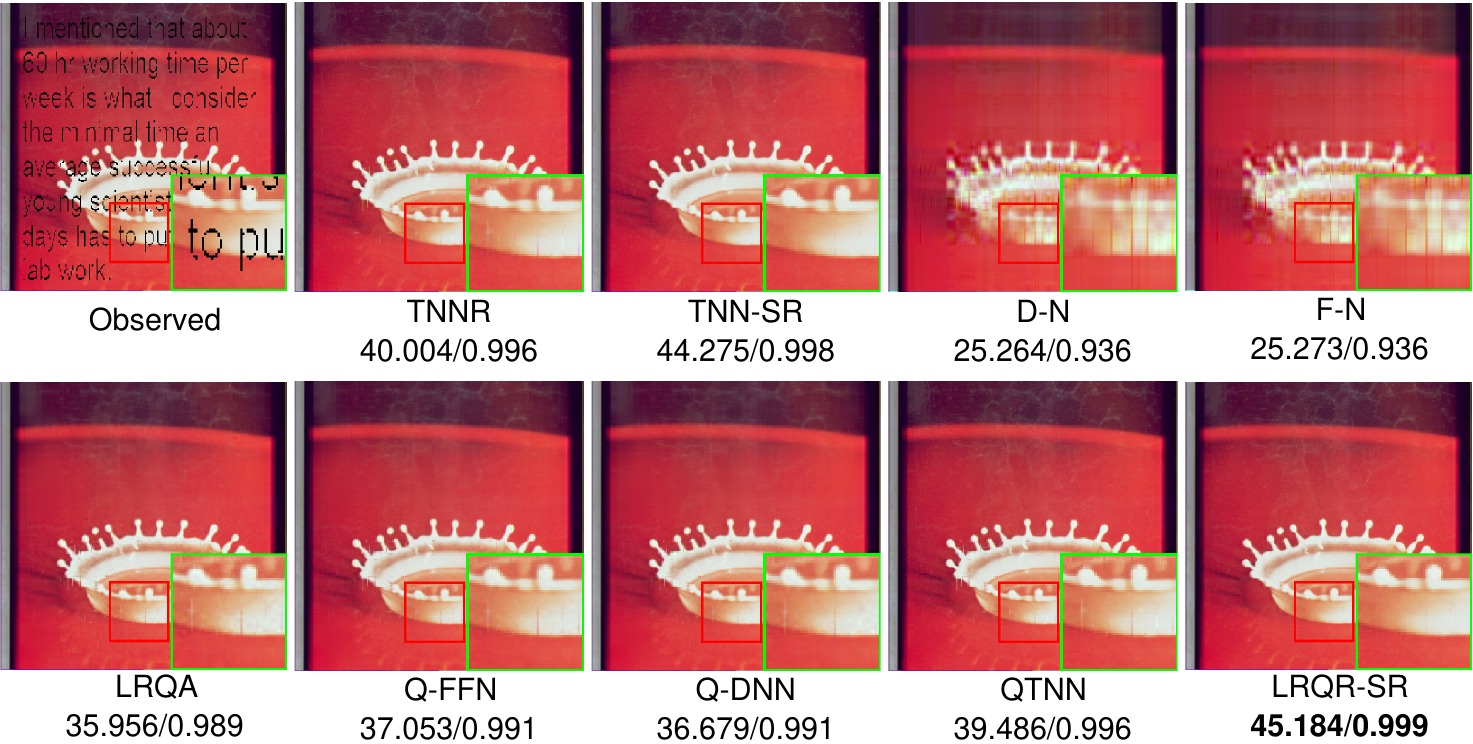}
	\caption{The recovered results of image $Splash$ by different methods under text mask.}
	\label{z8}
\end{figure}

Fig. \ref{z5}-\ref{z8} compares the visual results for all the competing completion approaches related to recovering $Tree, Beans, Flower, Splash$ under text mask. The image in the green box is a zoomed-in image of that in the red box. As shown, the recovered results from D-N and F-N are very blurry, especially where the content of the image is complex. It can also be observed from the zoomed-in portion of  Fig. \ref{z5} that the results for TNNR, LRQA, Q-FFN, and Q-DNN still leave some obvious artifacts in the blue area. Similar problems can be observed in Fig. \ref{z6}-\ref{z8}: In Fig. \ref{z6}, there are some obvious artifacts on the red beans in the zoomed-in portion, while in Fig. \ref{z7}, some vertical lines remain in the enlarged area after the restoration of the image. In Fig \ref{z8}, some further visible blemishes appear on the white part of the zoomed-in portion.  In comparison, the TNNR-SR and the proposed LRQR-SR approaches would obtain better performance, and though in general there is not much difference between the two approaches based on image observation, the corresponding PSNR and SSIM results show that the proposed method can restores the image technically more effectively.


\subsubsection{50 Images recovery to further demonstrate the effectiveness of our method}
\begin{figure*}[htbp]
	\centering
	\subfigure{
		\begin{minipage}{18cm}
			\centering
			\includegraphics[height = 3.5cm, width = 18 cm]{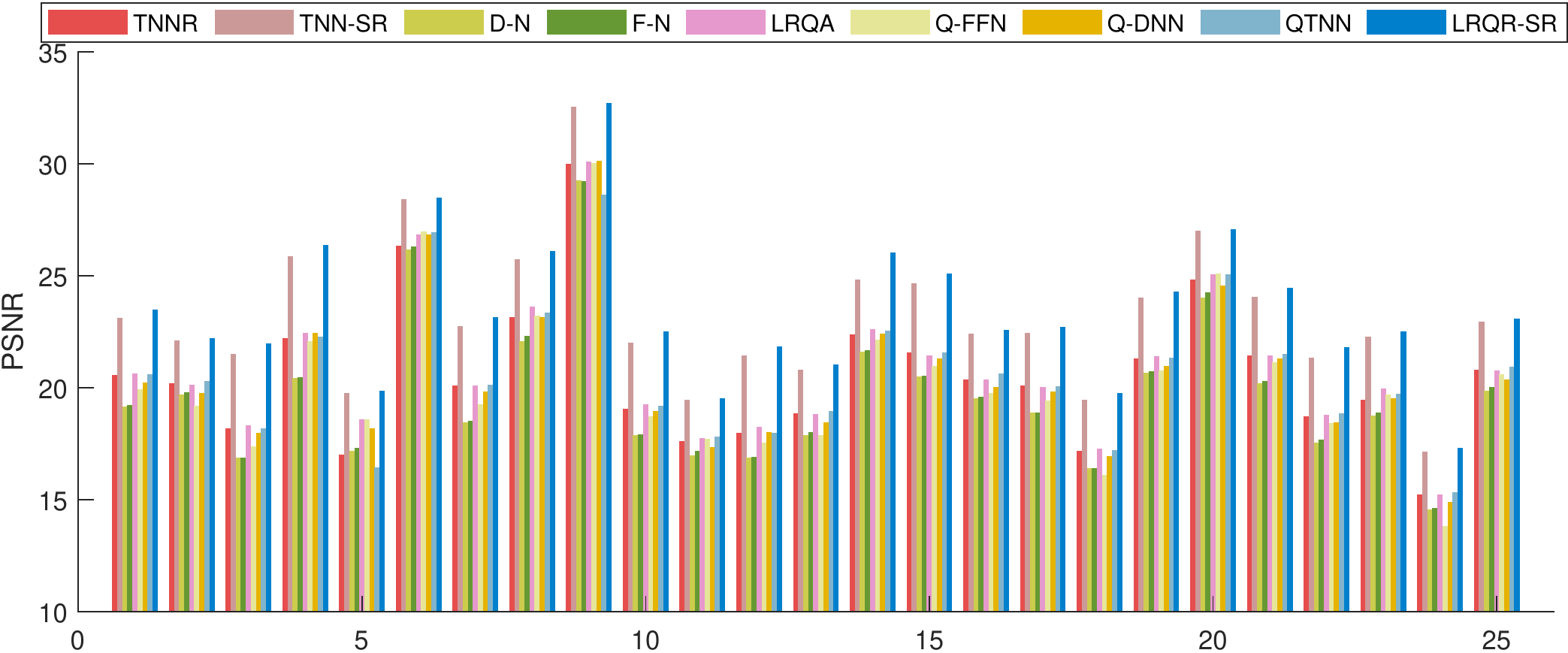}
		\end{minipage}
	}
	\subfigure{
		\begin{minipage}{18cm}
			\centering
			\includegraphics[height = 3.5cm, width = 18cm]{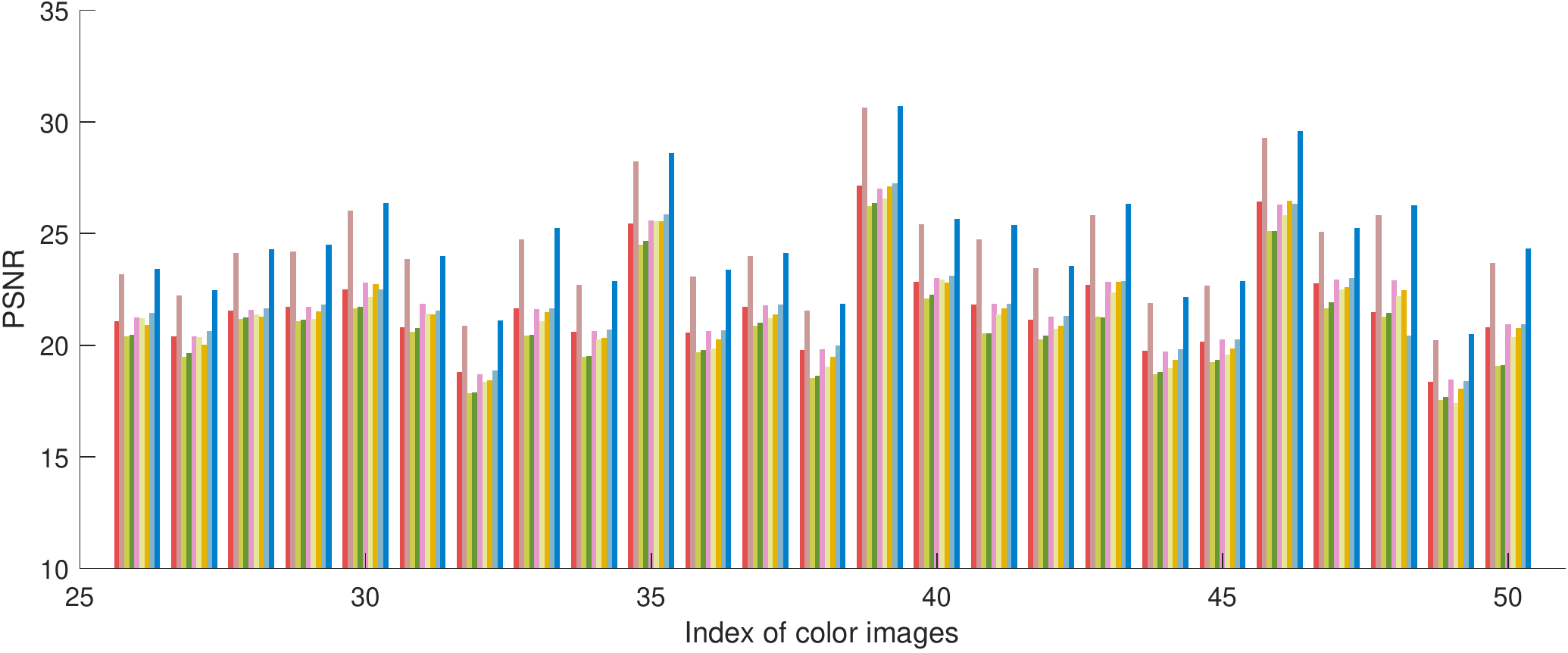}
		\end{minipage}
	}
	\caption{Comparison of  the PSNR results of different algorithms for recovering 50 color images selected from BSD (SR = 0.25).}
	\label{50p}
\end{figure*}

\begin{figure*}[htbp]
	\centering
	\captionsetup{font={footnotesize}}
	\subfigure{
		\begin{minipage}{18cm}
			\centering
			\includegraphics[height = 3.5cm, width = 18 cm]{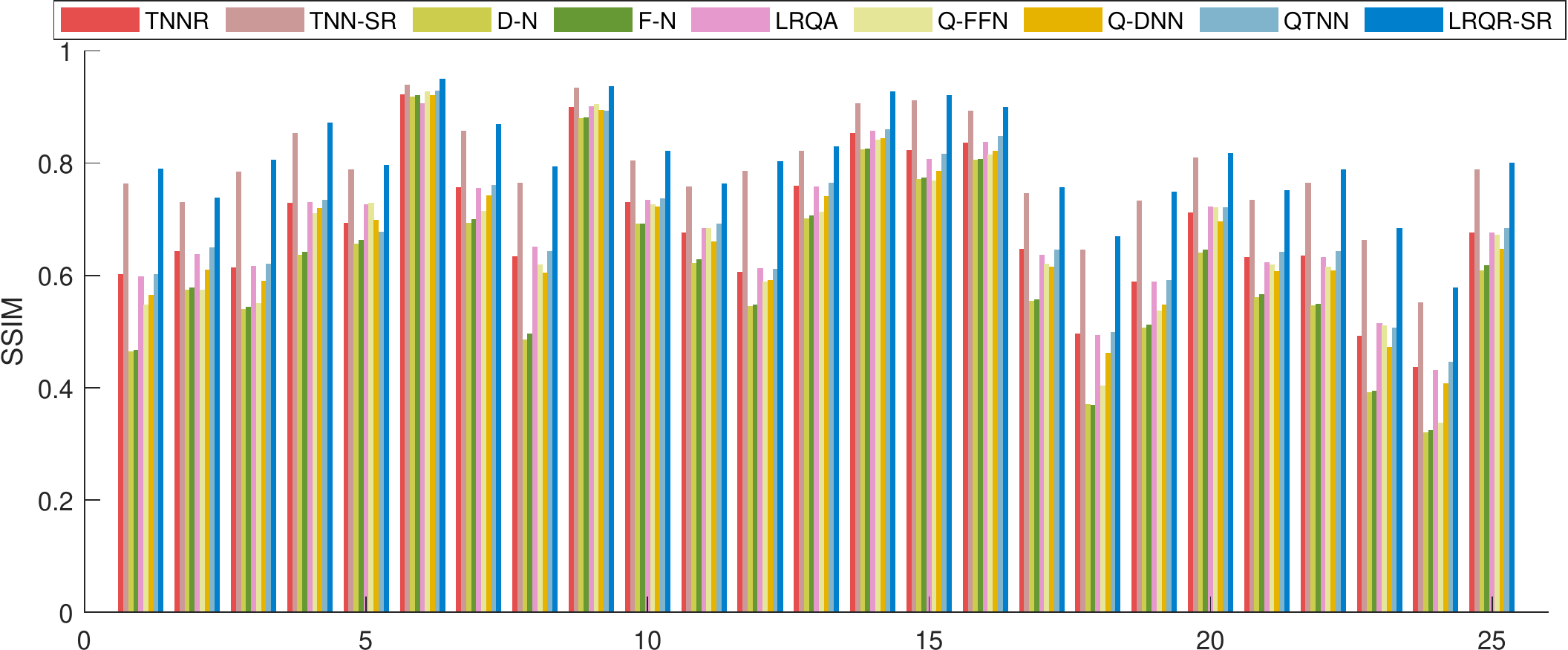}
		\end{minipage}
	}
	\subfigure{
		\begin{minipage}{18cm}
			\centering
			\includegraphics[height = 3.5cm, width = 18cm]{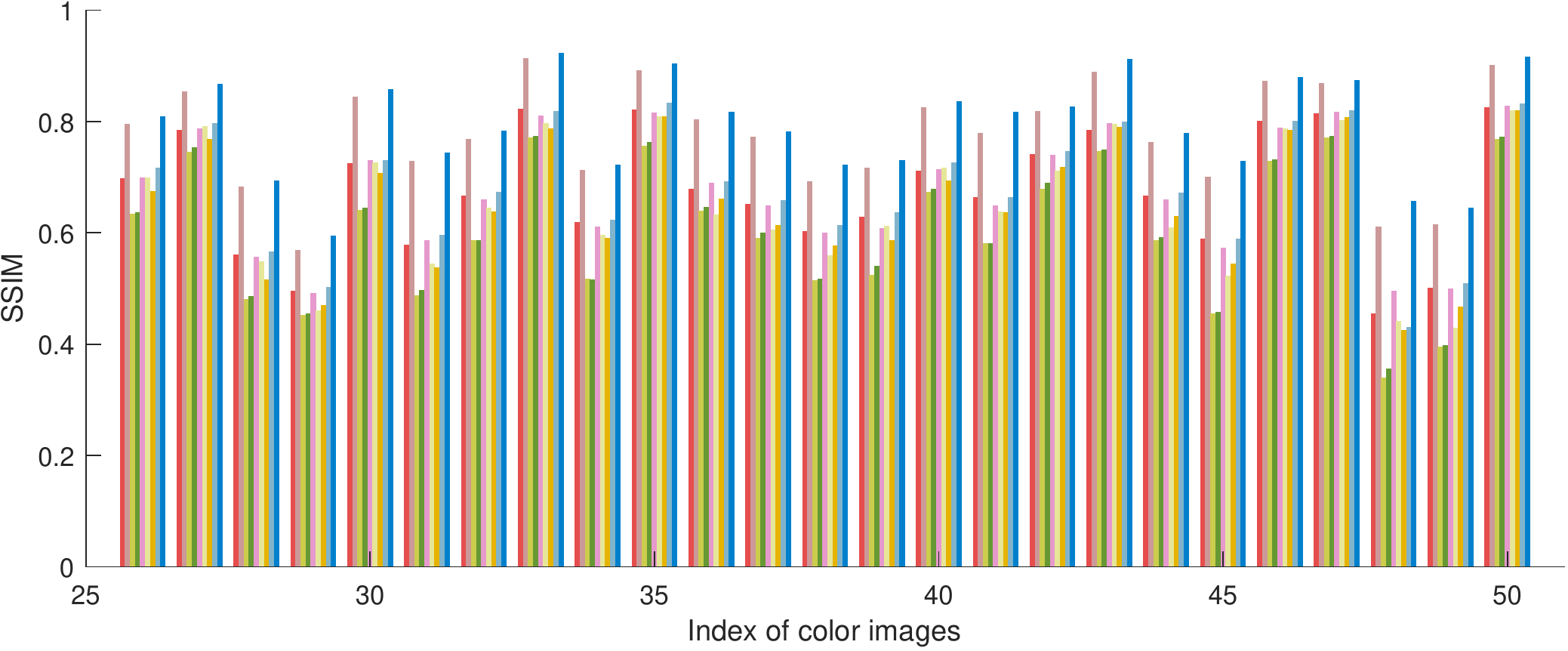}
		\end{minipage}
	}
	\caption{Comparison of the SSIM results of different algorithms for recovering 50 color images selected from BSD (SR = 0.25).}
	\label{50s}
\end{figure*}

A set of 50 images was subject to recovery to further demonstrate the effectiveness of the method. In this simulation, 50 images were randomly selected from BSD as a test sample, though in order to keep the parameters of the comparison algorithms as in the original set, these images were resized to $256\times 256\times 3$ throughout under random sample (SR=0.25). The corresponding PSNR and SSIM results are reported in Fig. \ref{50p} and Fig. \ref{50s}.

\subsection{Discussions} \label{d}
The simulation experiment results offer various items for discussion that can be summarised as follows:
\begin{itemize}
	\item The newly-developed LRQR-SR algorithm outperforms comparable existing algorithms both visually and numerically. The main reasons for this can be summarized in three points: this algorithm has been developed in the quaternion domain where the spatial structure information of color image is not destroyed;  the model uses QTNN to depict low-rankness, which helps preserve the information contained in the first few large singular values; and, finally, the $l_1$ norm is added to act as the regularization in the algorithm, helping to model the sparseness of the underlying quaternion matrix.
	\item When compared with matrix-based methods, TNNR, D-N, and F-N only depict low-rankness by means of modified NN or low-rank factorization. Hence, the recovery results are generally not satisfactory. However, as TNN-SR is based on both low-rankness and sparse priors, the recovered results are improved. In general, for matrix-based methods, the RGB channels of color images must be processed separately causing the recovered color images to potentially have details omitted.
	\item When compared with quaternion matrix-based methods, LRQA, Q-FFN, Q-DNN,and QTNN are limited by only utilizing low-rankness. low-rankness. For LRQR-SR, the $l_1$ norm is also incorporated to describe the sparse prior of the underlying quaternion matrix in the QDCT domain. Thus, LRQR-SR offers superior performance to the other comparable methods.
\end{itemize}
\section{Conclusion}
\label{C}
This paper proposed a novel low-rank quaternion recovery model incorporating sparse regularization that can be used to describe the connections of three dimensional structures to obtain better approximations. The proposed LRQR-SR method is based on the use of  QTNN to depict low-rankness, as well as taking advantage of the $l_1$ norm under QDCT to restrict the sparseness. A modified two-step ADMM framework was also adopted to optimize the model. The experimental results for actual color images illustrated the effectiveness of the resulting LRQR-SR, suggesting that, as the quaternion-based method can exploit correlations among color channels, it may be possible to combine this framework with deep learning methods in future work \cite{cao2021color}.

\section*{Acknowledgments}
This work is supported by the Science and Technology Planning Project of Guangzhou City, China (Grant No. 201907010043), University of Macau (MYRG2019-00039-FST), and the National Natural Science Foundation of China (Grant No. 62173308), and  the Natural Science
Foundation of Zhejiang Province of China (Grant No. LR20F030001 and D19A010003). The authors would like to express their heartfelt thanks to the health care workers on the front line of the fight against the COVID-19.  It is their dedication and sacrifice that provide a safe and stable research environment for people in this special era. 

{\appendix[Proof of  Theorem \ref{th3}]
 According to \cite{DBLP:journals/tip/ChenXZ20}, we need to prove
	\begin{equation}\label{p1}
\min\limits_{\dot{\mathbf{X}}}\parallel\dot{\mathbf{Y}}-\dot{\mathbf{X}}\parallel_F^2+\lambda \parallel\dot{\mathbf{X}}\parallel_1
	\end{equation}
  has one unique optimal solution $\dot{\mathbf{X}}_{\star}$ and $\dot{\mathbf{X}}_{\star}$ equals to $\dot{\mathbf{X}}_{opt}$ defined in \eqref{eqth3}. 
  
  \begin{proof}
 It can be observed that two terms in \eqref{p1} are convex, hence, \eqref{p1} has one unique optimal solution. Based on the rules of quaternion matrix derivatives in \cite{DBLP:journals/tsp/XuM15}, $\dot{\mathbf{X}}_{\star}$ must satisfy the following formula:
 \begin{equation}\label{p2}
 \dot{0}\in\dot{\mathbf{X}}_{\star}-\dot{\mathbf{Y}}+2\lambda\partial\parallel\dot{\mathbf{X}}_{\star}\parallel_1,
 \end{equation}
where $\partial\parallel\dot{\mathbf{X}}_{\star}\parallel_1$ represents the subgradient of $\parallel\dot{\mathbf{X}}_{\star}\parallel_1$. Following \cite{DBLP:journals/nla/JiaNS19}, the subgradient of the $l_1$ norm at $\dot{\mathbf{X}}_{\star}$ is given by 
 \begin{equation}\label{p3}
\partial\parallel\dot{\mathbf{X}}_{\star}\parallel_1=\{\dot{\mathbf{G}}\in\mathbb{H}^{M \times N}: \dot{\mathbf{G}}=direc(\dot{\mathbf{X}}_{\star})+\dot{\mathbf{F}}, \parallel\dot{\mathbf{F}}\parallel_\infty\leq1\},
\end{equation}
where $direc(\dot{\mathbf{X}}_{\star})$ is a $M \times N$ matrix with the entries computed by $[\frac{x_{ij}}{\mid x_{ij}\mid}]_{M \times N}$. 
Then, $\dot{\mathbf{X}}_{opt}$ need to be proved to satisfy \eqref{p2}.\\
When $\dot{y}>2\lambda$, $\dot{x}>0$, then\\ $\dot{y}-\dot{x}=\dot{y}-\frac{\dot{y}}{\mid \dot{y}\mid}\max\{\mid\dot{y}\mid-2\lambda,0\}=\dot{y}-(\dot{y}-2\lambda)=2\lambda$.\\
When $-2\lambda\leqslant\dot{y}\leqslant2\lambda$, $\dot{x}=0$, then\\ $\dot{y}-\dot{x}=\dot{y}$. Let $\dot{\mathbf{F}}=\frac{1}{2\lambda}\dot{\mathbf{Y}}$, then we have $\parallel\dot{\mathbf{F}}\parallel_\infty\leqslant1$.\\
When $\dot{y}<-2\lambda$, $\dot{x}<0$, then\\ 
$\dot{y}-\dot{x}=\dot{y}+(-\dot{y}-2\lambda)=-2\lambda=2\lambda(-1)$.\\
Based on the above discussions, we can obtain that $\dot{0}\in\dot{\mathbf{X}}_{opt}-\dot{\mathbf{Y}}+2\lambda\partial\parallel\dot{\mathbf{X}}_{opt}\parallel_1$, which means that $\dot{\mathbf{X}}_{opt}=\dot{\mathbf{X}}_{\star}$.
\end{proof}


\bibliographystyle{IEEEtran}
\bibliography{IEEEabrv, mybibfile}

\end{document}